\documentclass[runningheads]{llncs}
\usepackage{graphicx}
\usepackage[normalem]{ulem}
\usepackage{amsmath,amssymb} 
\usepackage{color}
\usepackage[width=122mm,left=12mm,paperwidth=146mm,height=193mm,top=12mm,paperheight=217mm]{geometry}

\usepackage[pagebackref=true,breaklinks=true,letterpaper=true,colorlinks,bookmarks=false]{hyperref}
\usepackage{times}
\usepackage{epsfig}
\usepackage{graphicx}
\usepackage{amsmath}
\usepackage{amssymb}

\usepackage{times}
\usepackage{epsfig}
\usepackage{amsmath}
\usepackage{amssymb}
\usepackage{multirow}
\usepackage{wrapfig}
\usepackage{caption}
\usepackage{soul}
\usepackage{booktabs}

\begin{document}

\newcommand{\old}[1]{{\color{red}\st{}}}
\newcommand{\new}[1]{{#1}}
\newcommand{\huaizu}[1]{{\color{red}HJ: #1}}
\newcommand{\gs}[1]{\textcolor{magenta}{#1}}
\newcommand\etal{\textit{et al.}}
\newcommand\ie{\textit{i.e.}}
\newcommand\eg{\textit{e.g.}}

\newcommand{\first}[1]{\textbf{#1}}
\newcommand{\second}[1]{\textbf{#1}}
\newcommand{\third}[1]{#1}

\addtolength{\parskip}{-0.5pt}

\pagestyle{headings}
\mainmatter
\def\ECCV18SubNumber{67}  

\title{Self-Supervised Relative Depth Learning for\\ Urban Scene Understanding} 


\authorrunning{Jiang, Learned-Miller, Larsson, Maire and Shakhnarovich}


\author{Huaizu Jiang$^1$,~~~~~Erik Learned-Miller$^1$\\
Gustav Larsson$^2$,~~~Michael Maire$^3$,~~~Greg Shakhnarovich$^3$}
\institute{$^1$UMass Amherst~~~~~$^2$University of Chicago~~~~~$^3$TTI-Chicago}

\maketitle

\begin{abstract}
As an agent moves through the world, the apparent motion of scene elements is (usually) inversely proportional to their depth.\footnote{\label{note1}Strictly speaking, this statement is true only after one has compensated for camera rotation, individual object motion, and image position. We address these issues in the paper.} It is natural for a learning agent to associate image patterns  with the magnitude of their displacement over time: as the agent moves, faraway mountains don't move much;  nearby trees move a lot. This natural relationship between the appearance of objects and their motion is a rich source of information about the world. 
In this work, we start by training a deep network, using {fully automatic supervision}, to predict relative scene depth from \emph{single images}. The relative depth training images are automatically derived from simple videos of cars moving through a scene, using recent motion segmentation techniques, and no human-provided labels.  This {\it proxy} task of predicting relative depth from a single image induces features in the network that result in large improvements in a set of downstream tasks including semantic segmentation, joint road segmentation and car detection, and monocular (absolute) depth estimation, over a network trained from scratch. The improvement on the semantic segmentation task is greater than those produced by any other automatically supervised\footnote{We refer to our method as {\it automatically supervised} or {\it self-supervised} when we want to emphasize that training data was obtained automatically. At other times, we refer to it as {\it unsupervised} to emphasize that it uses no human supervision.} methods. Moreover, for monocular depth estimation, our unsupervised pre-training method even outperforms supervised pre-training with ImageNet. In addition, we demonstrate benefits from learning to predict (again, completely unsupervised) relative depth in the specific videos associated with various downstream tasks (\eg, KITTI). We adapt to the specific scenes in those tasks in an unsupervised manner to improve performance. In summary, for semantic segmentation, we present state-of-the-art results among methods that do not use supervised pre-training, and we even exceed the performance of supervised ImageNet pre-trained models for monocular depth estimation, achieving results that are comparable with state-of-the-art methods. 
\end{abstract}

\section{Introduction}
How does a newborn agent learn about the world? When an animal (or robot) moves, its visual system is exposed to a shower of information. Usually, the speed with which something moves in the image is inversely proportional to its depth.\footnotemark[1]
As an agent continues to experience visual stimuli under its own motion, it is natural for it to form associations between the appearance of objects and their relative motion in the image. For example, an agent may learn that objects that look like mountains typically don't move in the image (or change appearance much) as the agent moves. Objects like nearby buildings and bushes, however, appear to move rapidly in the image as the agent changes position relative to them. This continuous pairing of images with motion acts as a kind of automatic supervision that could eventually allow an agent both to understand the depth of objects and to group pixels into objects by this predicted depth. Thus, by moving through the world, an agent may learn to predict properties (such as depth) of {\it static} scenes.

\begin{figure*}[t]
\renewcommand{\arraystretch}{0.6}
\renewcommand{\tabcolsep}{0.8pt}
\centering
\begin{tabular}{ccccc}
	\includegraphics[height=0.09\linewidth]{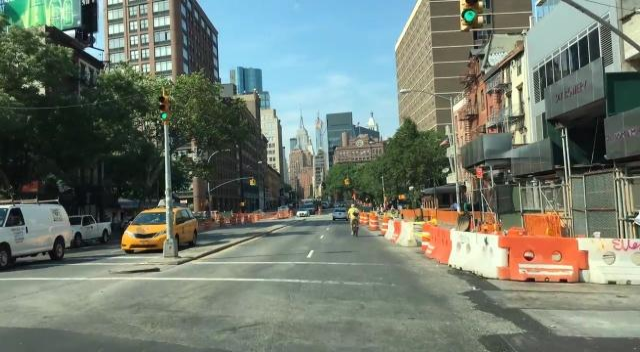} & 
	\includegraphics[height=0.09\linewidth]{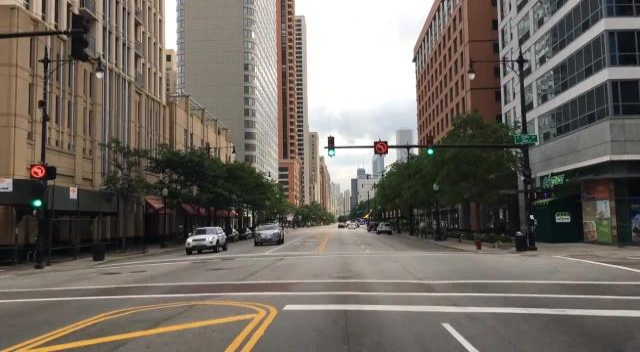} &
    \includegraphics[height=0.09\linewidth]{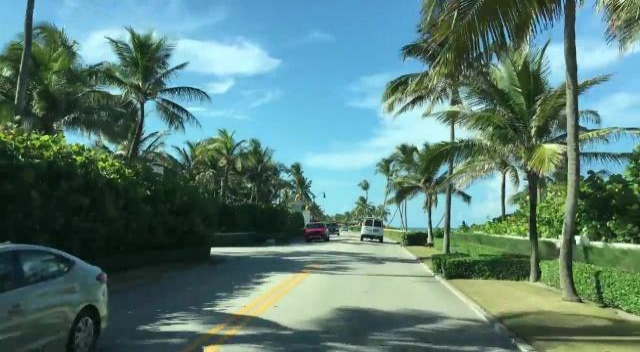} &
    \includegraphics[height=0.09\linewidth]{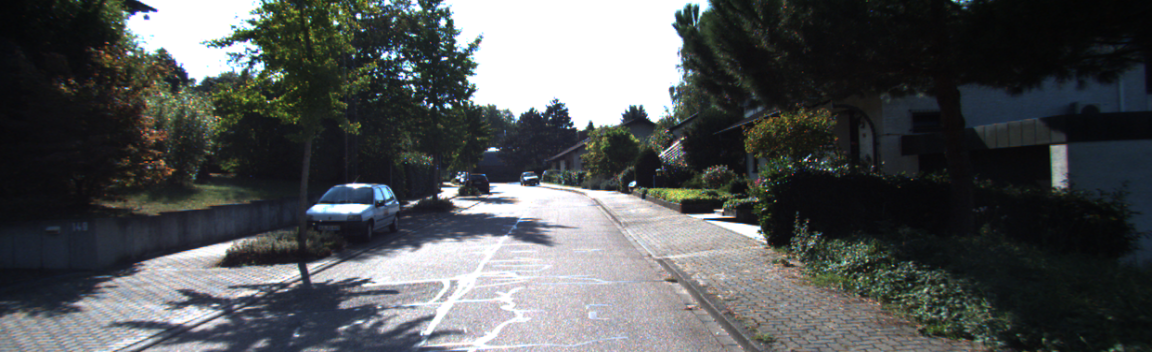} &
    \includegraphics[height=0.09\linewidth]{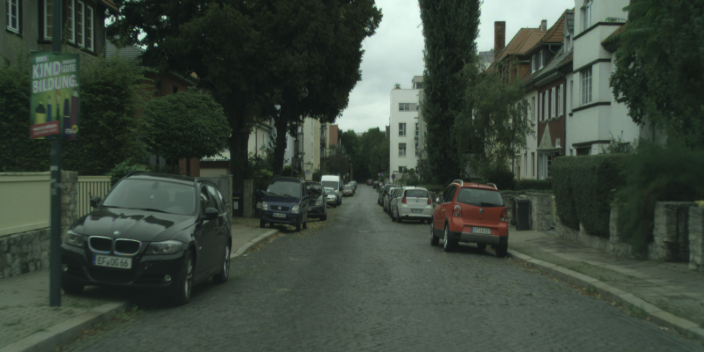} \\
    \includegraphics[height=0.09\linewidth]{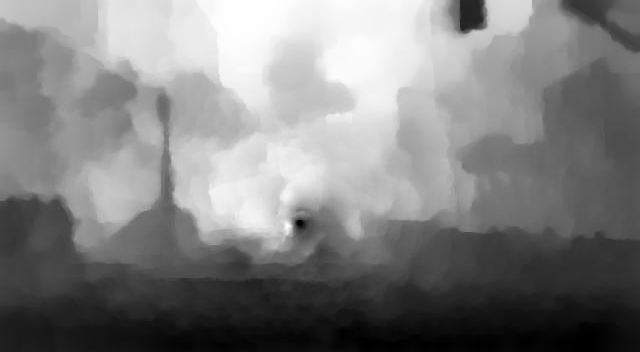} & 
	\includegraphics[height=0.09\linewidth]{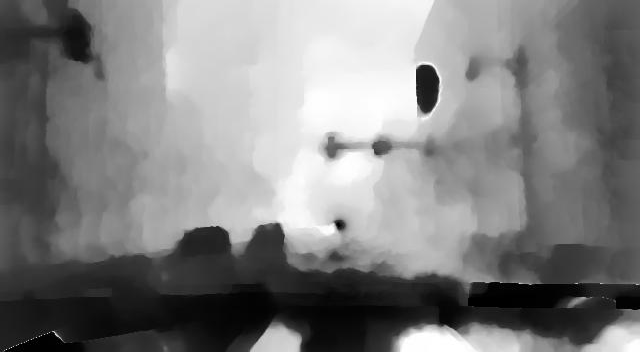} &
    \includegraphics[height=0.09\linewidth]{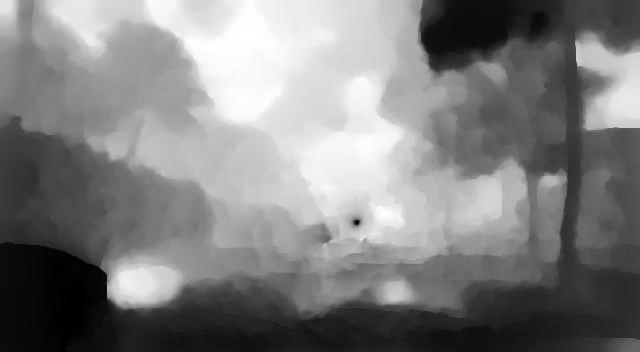} &
    \includegraphics[height=0.09\linewidth]{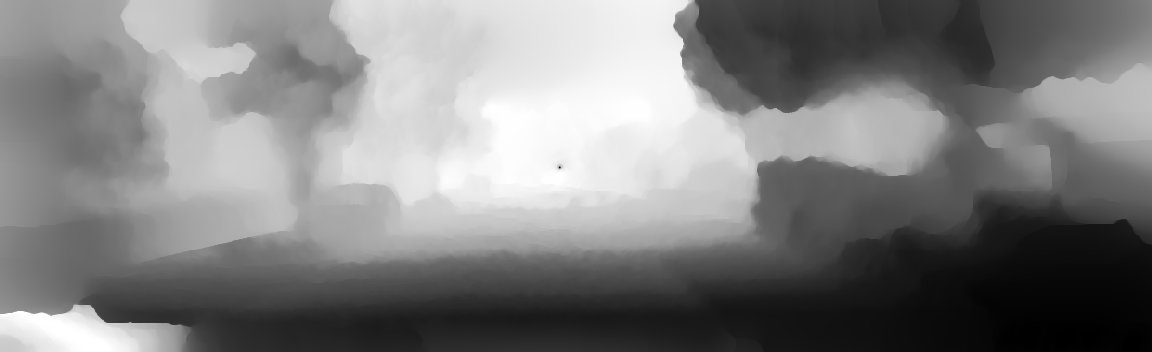} &
    \includegraphics[height=0.09\linewidth]{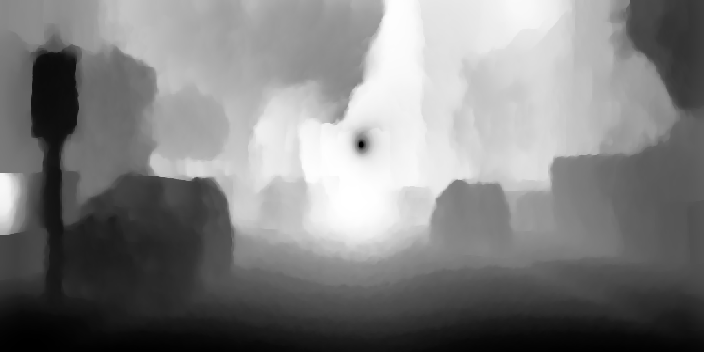} \\
    \includegraphics[height=0.09\linewidth]{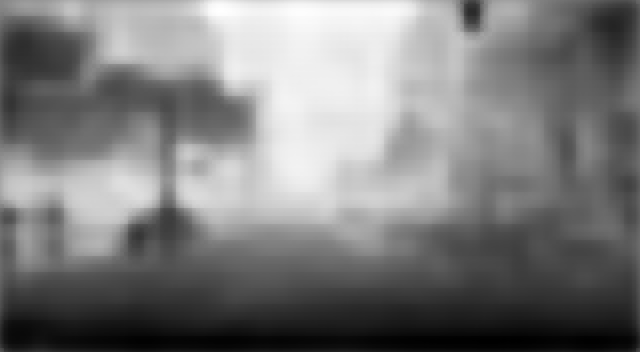} & 
	\includegraphics[height=0.09\linewidth]{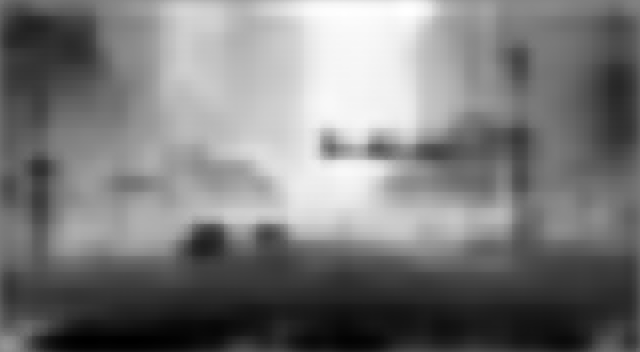} &
    \includegraphics[height=0.09\linewidth]{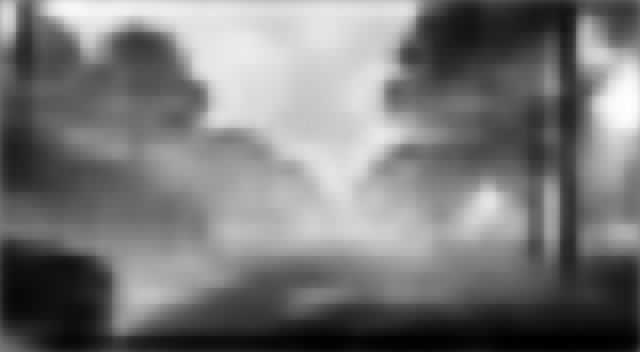} &
    \includegraphics[height=0.09\linewidth]{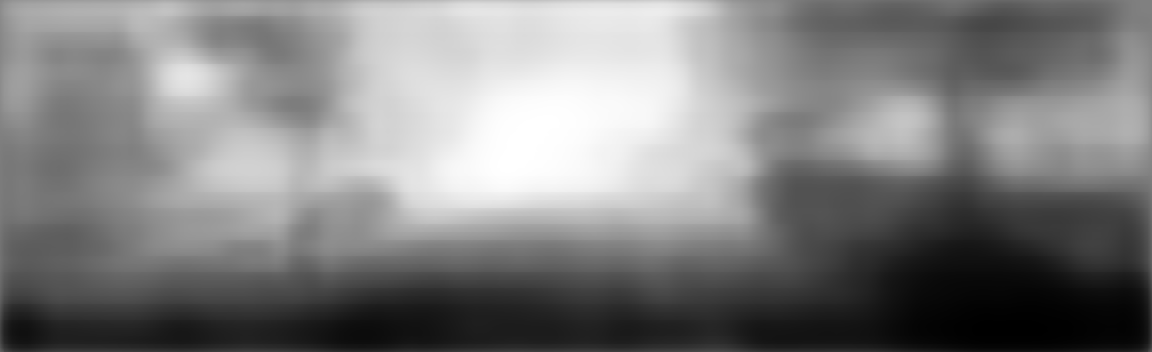} &
    \includegraphics[height=0.09\linewidth]{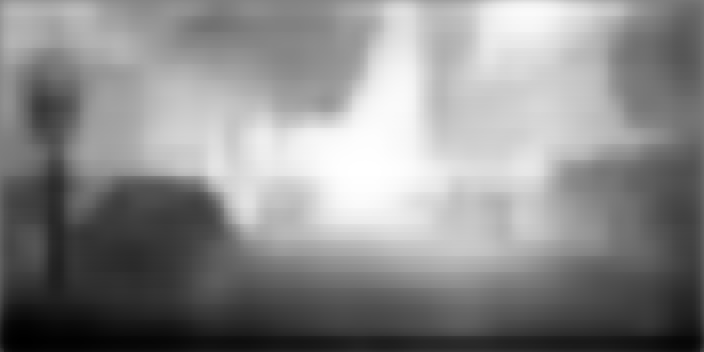} \\
    (a) & (b) & (c) & (d) & (e) \\
\end{tabular}
\caption{Sample frames from collected videos and their corresponding relative depth maps, where brightness encodes relative depth (the brighter the further). From top to bottom: input image, relative depth image computed using Eq.(\ref{eq:depth_recovery}), and predicted (relative) depth maps using our trained VGG16 FCN8s~\cite{simonyan14very,shelhamer17fully}. There is often a black blob around the center of the image, a singularity in depth estimation caused by the focus of expansion. (a)(b)(c): images from the CityDriving dataset, (d): images from the KITTI dataset, and (e): images from the CityScapes dataset.}
\label{fig:samples_im_depth_pairs}
\end{figure*}

A flurry of recent work has shown that {\it proxy  tasks} (also known as {\it pretext} or {\it surrogate tasks}) such as colorization~\cite{larsson2017colorproxy,zhang2016colorful}, jigsaw puzzles~\cite{noroozi2016jigsaw}, and others~\cite{wang2015unsupervised,agrawal15learning,pathak2017move,misra2016unsupervised,pathak2016context,gao16object,owens2016ambient,lee17unsupervised}, can induce features in a neural network that provide strong pre-training for subsequent tasks. In this paper, we introduce a new proxy task: estimation of relative depth from a single image. We show that a network that has been pre-trained, without human supervision, to predict relative scene depth provides a powerful starting point from which to fine-tune models for a variety of urban scene understanding tasks.  Not only does this automatically supervised starting point outperform all other proxy task pre-training methods. For monocular depth understanding, it even performs better than the heavily supervised ImageNet pre-training, yielding results that are comparable with state-of-the-art methods.

To estimate relative scene depths without human supervision, we resort to a recent motion segmentation technique~\cite{bideau16its} to estimate relative depth from geometric constraints between scene's motion field and camera motion. We apply it to simple, publicly available YouTube videos taken from moving cars. Since this technique estimates depth {\em up to an unknown scale factor}, we compute {\em relative depth} of the scene during the pre-training phase, where each pixel's value is in the range of $[0, 1]$ denoting its depth percentile over the entire image.\footnote{Later, we will fine-tune networks to produce absolute depths.} 

Unlike work that analyzes video paired with additional information about direction of motion~\cite{jayaraman15learning}, our agent learns from  ``raw egomotion'' video recorded from cars moving through the world. Unlike methods that require videos of moving objects~\cite{pathak2017move}, we neither depend on, nor are disrupted by, moving objects in the video. Once we have relative depth estimates for these video images, we train a deep network to predict the relative depth of each pixel from a {\em single image}, \ie, to predict the relative depth {\em without} the benefit of motion. One might expect such a network to learn that an image patch that looks like a house and spans 20 pixels of an image (about 100 meters away) is significantly further away than a pedestrian that spans 100 image pixels (perhaps 10 meters away). 
Figure \ref{fig:samples_im_depth_pairs} illustrates this prediction task and shows sample results obtained using a standard convolutional neural network (CNN) in this setting.  For example, in the leftmost image of  Fig.~\ref{fig:samples_im_depth_pairs}, an otherwise unremarkable traffic-light pole is clearly highlighted  by its relative depth profile, which stands out from the background.  Our hypothesis is that to excel at relative depth estimation, the CNN will benefit by learning to recognize such structures. 

The goal of our work is to show that pre-training a network to do relative depth prediction is a powerful  proxy task for learning visual representations.  In particular, we show that a network pre-trained for relative depth prediction (from automatically generated training data) improves training for downstream tasks including semantic segmentation, joint semantic reasoning of road segmentation and car detection, and monocular (absolute) depth estimation.
We obtain significant performance gains on urban scene understanding benchmarks such as KITTI~\cite{Geiger2012CVPR,Geiger2013IJRR} and CityScapes~\cite{Cordts2016Cityscapes}, compared to training a segmentation model from scratch. Compared to nine other proxy tasks for pre-training, our proxy task consistently provides the highest gains when used for pre-training. In fact, our performance on semantic segmentation and joint semantic reasoning tasks comes close to that of equivalent architectures pre-trained with ImageNet~\cite{deng2009imagenet}, a massive labeled dataset. Finally, for the monocular (absolute) depth estimation, \emph{our pre-trained model achieves better performance than an ImageNet pre-trained model}, using both VGG16~\cite{simonyan14very} and ResNet50~\cite{he16deep} architectures.

As a final application, we show how our proxy task can be used for {\it domain adaptation}.
One might assume that the more similar the domain of unlabeled visual data used for the proxy task (here, three urban scene understanding tasks) is to the domain in which the eventual semantic task is defined (here, semantic segmentation), the better the representation learned by pre-training.
This observation allows us to go beyond simple pre-training, and effectively provide a domain adaptation mechanism. By adapting (fine-tuning) a relative depth prediction model to targets obtained from unlabeled data in a novel domain (say, driving in a new city) we can improve the underlying representation, priming it for better performance on a semantic task (\eg, segmentation) trained with a small labeled dataset from this new domain. In experiments, we show that pre-training on unlabeled videos from a target city, absent any labeled data from that city, consistently improves all urban scene understanding tasks.

In total, our work advances two pathways for integrating unlabeled data with visual learning.  
\begin{itemize}
\item We propose a novel proxy task for self-supervised learning of visual representations; it is based on learning to predict relative depth, inferred from unlabeled videos.  This unsupervised pre-training leads to better results over all other proxy tasks on the semantic segmentation task, and even outperforms supervised ImageNet pre-training for absolute depth estimation.  
\item We show that our task can be used to drive domain adaptation.  Experiments demonstrate its utility in scene understanding tasks for street scenes in a novel city. Our adapted model achieves results that are competitive with state-of-the-art methods (including those that use \emph{large supervised pre-training}) on the KITTI depth estimation benchmark.
\end{itemize}
Such methods of extracting knowledge from unlabeled data are likely to be increasingly important as computer vision scales to real-world applications; here massive dataset size can outpace herculean annotation efforts.

\section{Related Work}\label{sec:related}


\noindent\textbf{Self-supervised learning.} The idea of formulating supervised prediction tasks on unlabeled data has been
leveraged for both images and videos. The idea, often called self-supervision, is most typically realized by removing part of the input and then
training a network to predict it.  This can take the form of deleting a spatial
region and trying to inpaint it~\cite{pathak2016context}, draining an image of color
and trying to colorize
it~\cite{larsson2016learning,zhang2016colorful,larsson2017colorproxy}, or
removing the final frame in a sequence and trying to hallucinate
it~\cite{ranzato2014video,srivastava2015video,mathieu2015deep,vondrick2016generating,xue2016visual,lotter2017prednet}.
Generative Adversarial Networks, used for inpainting and several future frame
prediction methods, can also be used to generate realistic-looking
samples from scratch. This has found secondary utility for unsupervised
representation learning~\cite{radford2016unsupervised,springenberg2016unsupervised,donahue2017adversarial}.
Another strategy is to extract patches and try to predict their spatial or temporal relationship. In
images, this has been done for pairs of patches~\cite{doersch2015unsupervised}
or for 3-by-3 jigsaw puzzles~\cite{noroozi2016jigsaw}. In videos, it can be done by predicting the temporal ordering of frames~\cite{misra2016unsupervised,lee17unsupervised}. The correlation of frames in video is also a rich source of self-supervised learning signals. The assumption that close-by frames are more similar than far apart frames can be used to train embeddings on pairs~\cite{mobahi2009deep,isola2015learning,jayaraman2015slow} or triplets~\cite{wang2015unsupervised} of frames. A related idea that the representation of interesting objects should change slowly through time dates back to Slow Feature Analysis~\cite{wiskott2002slow}. 

The works most closely related to ours may be~\cite{jayaraman15learning,agrawal15learning,pathak2017move}, which aim to learn useful visual representations from unlabeled videos as well. Jayaraman \& Grauman~\cite{jayaraman15learning} learn a representation equivariant to ego-motion transformations, using ideas from metric learning.
Agrawal~\etal~\cite{agrawal15learning} concurrently developed a similar
method that uses the ego-motion directly as the prediction target as opposed
to as input to an equivariant transformation. Both of these works assume
knowledge of the agent's own motor actions, which limits their evaluation in sample size
due to lack of publicly available data. In our work, the ego-motion is inferred
through optical flow, which means we can leverage large sources of crowd-sourced
data, such as YouTube videos. Pathak~\etal~\cite{pathak2017move} use optical flow and a
graph-based algorithm to produce unsupervised segmentation maps. A network is then
trained to approximate these maps, driving representation learning. The
reliance on moving objects, as opposed to a moving agent, could make it harder
to collect good data. Using a method based on ego-motion, the agent can promote
its own representation, learning simply by moving, instead of having to find
objects that move.

There is also work on using multi-modal sensory input as a source of supervision. Owens~\etal~\cite{owens2016ambient} predict statistics of ambient sounds in videos. Beyond studying a single source of self-supervision, combining multiple self-supervision sources is increasingly popular. In~\cite{doersch17multi}, a set of self-supervision tasks are integrated via a multi-task setting. Wang~\etal~\cite{wang17transitive} propose to combine instance-level as well as category-level self-supervision. Both~\cite{doersch17multi,wang17transitive} achieve better performance than a single model.

\noindent\textbf{Unsupervised learning of monocular depth estimation.} A single-image depth predictor can be trained from raw stereo images, by warping the right image with a depth map predicted from the left image and train it to reconstruct the left image~\cite{garg16unsupervised,godard16unsupervised}. 
This idea was extended in recent work to support fully self-supervised training on regular video, by predicting both depth and camera pose difference for pairs of nearby frames~\cite{zhou17unsupervised,vijayanarasimhan17sfmnet}. 

Although~\cite{zhou17unsupervised,vijayanarasimhan17sfmnet} are closely related to our work in the sense of unsupervised (or self-supervised) learning depth and ego-motion from unlabeled videos, our work differs from them in two ways. First, neither of these two works emphasizes more general-purpose feature learning. Second, neither of them demonstrates their scalability to large-scale YouTube videos. \cite{zhou17unsupervised} requires intrinsic camera parameters that are not available for most YouTube videos; our approach relies on optical flow only. \cite{vijayanarasimhan17sfmnet} only reports experimental results on standard benchmark datasets, whose scale is an order of magnitude smaller than videos we use. It is unclear whether the heuristics (\eg, the manually set camera intrinsic parameters, number of motion clusters) are robust to YouTube videos in the wild.

\section{Inducing features by learning to estimate relative depth}
As a proxy task, our goal is to induce a feature representation $f(I)$ of an RGB image $I(x,y)$ by predicting its depth image $z(x,y)$, where the representation $f(I)$ could be transferred to other downstream tasks (\eg, semantic segmentation) with  fine-tuning. In section~\ref{sec:self_supervised_depth}, we introduce technical details of gathering images and corresponding depth maps. In section~\ref{sec:network_training}, we provide details of training CNNs to learn the feature representation $f(I)$.

\subsection{Self-Supervised Relative Depth} 
\label{sec:self_supervised_depth}
As described above, we automatically produce depth images for video frames by analyzing the motion of pre-existing videos. In our experiments, we used three sets of videos: YouTube videos, videos from the KITTI database~\cite{Geiger2012CVPR,Geiger2013IJRR}, and videos from the CityScapes database~\cite{Cordts2016Cityscapes}. The YouTube videos consist of 135 videos taken from moving cars in major U.S.~cities.\footnote{They are crawled from a YouTube playlist, taking less than an hour.} We call this dataset CityDriving. The stability of the camera in these videos makes them relatively easy for the depth estimation procedure. Some of the videos are extremely long, lasting several hours. The CityDriving dataset features a large number of man-made structures, pedestrians and cars. Following~\cite{li16unsupervised}, we only keep two consecutive frames if they have moderate motion (\ie, neither too slow nor too fast). To eliminate near duplicate frames, two consecutive depth maps must be at least 2 frames apart. We keep only the first one of two consecutive frames and the computed depth image. In total, we gathered 1.1M pairs of RGB images and their corresponding depth maps, where the typical resolution is $640 \times 360$. Similarly, we collect 30K and 24K pairs of RGB images and their relative depth maps for CityScapes and KITTI, respectively.

Denote the instantaneous coordinates of a point $P$ in the environment by $(X, Y, Z)^T$, and the translational velocity of the camera in the environment by $(U, V, W)^T$. Let the motion field component (idealized optical flow) of the point $P$ (in the image plane) be $(u, v)$, corresponding to the horizontal and vertical image motion, respectively. The motion field can be written as the sum of translation and rotation components\footnote{Any motion in the image is due to the relative motion of a world point and the camera. This addresses motion of  the object, the camera, or both.}
\begin{align}
u=u_t + u_r,~~~~~v=v_t + v_r,
\end{align}
where the subscript $t$ and $r$ denote translation and rotation, respectively. According to the geometry of perspective projection~\cite{Horn_1986}, the following equations hold if the motion of the camera is purely translational,
\begin{align}
u_t=\frac{-U+xW}{Z},~~~~~v_t=\frac{-V+yW}{Z},
\end{align}
where $x$ and $y$ are the coordinates of the point $P$ in the image plane (the origin is at the image center). 

\begin{figure}[t]
\centering
\renewcommand{\arraystretch}{0.6}
\renewcommand{\tabcolsep}{0.8pt}
\begin{tabular}{cccc}
\includegraphics[width=0.25\linewidth]{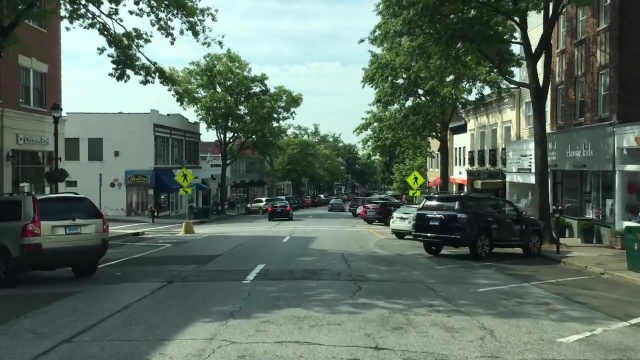} & 
\includegraphics[width=0.25\linewidth]{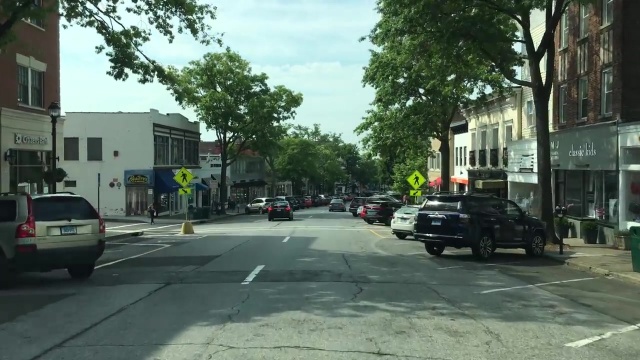} & 
\includegraphics[width=0.25\linewidth]{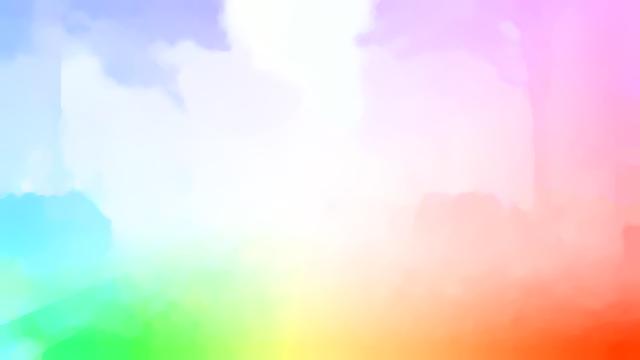} & 
\includegraphics[width=0.25\linewidth]{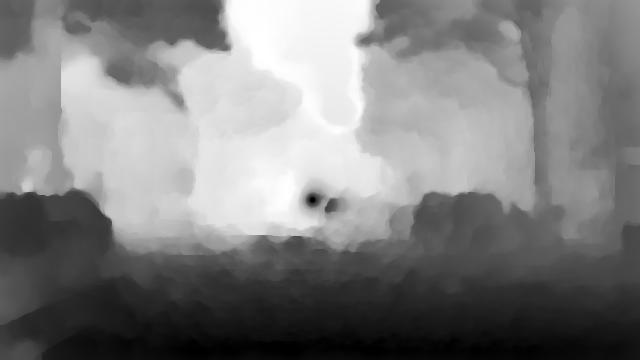} \\
\includegraphics[width=0.25\linewidth]{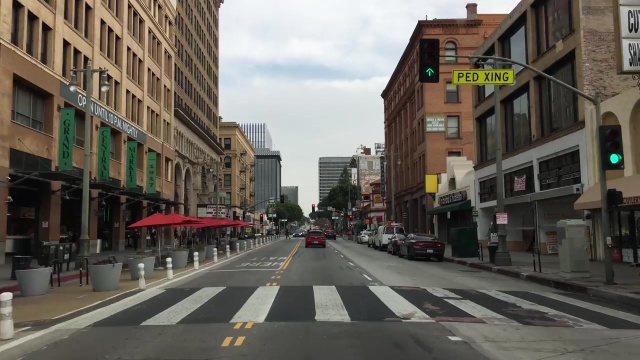} & 
\includegraphics[width=0.25\linewidth]{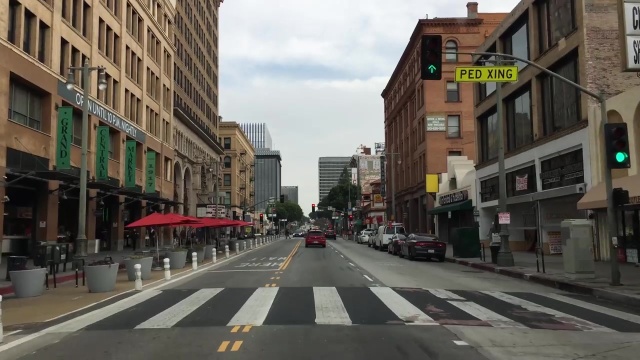} & 
\includegraphics[width=0.25\linewidth]{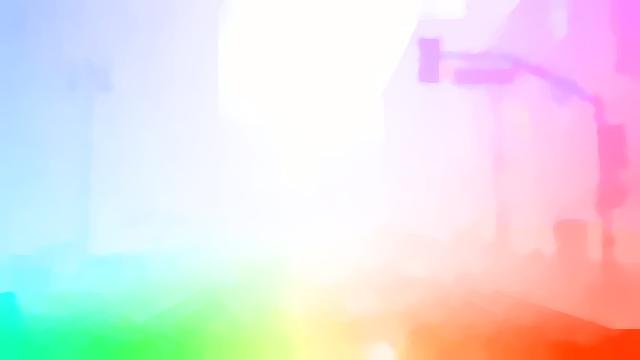} & 
\includegraphics[width=0.25\linewidth]{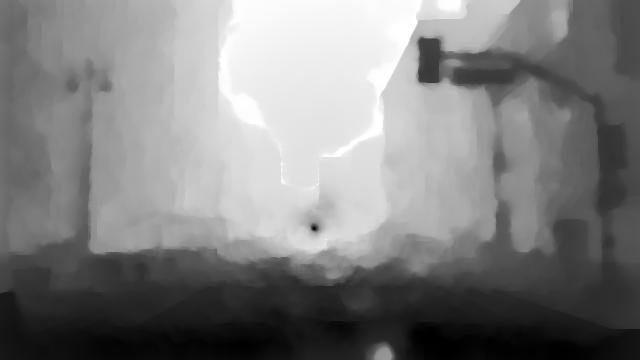} \\
\includegraphics[width=0.25\linewidth]{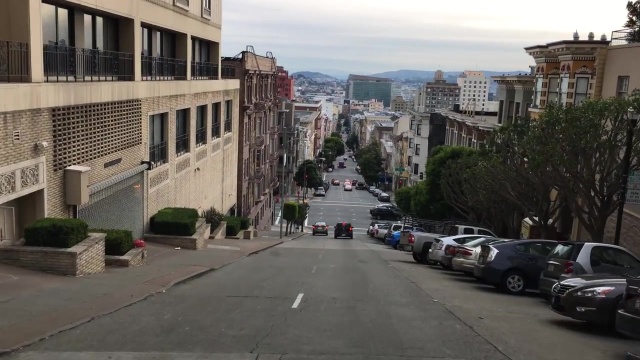} & 
\includegraphics[width=0.25\linewidth]{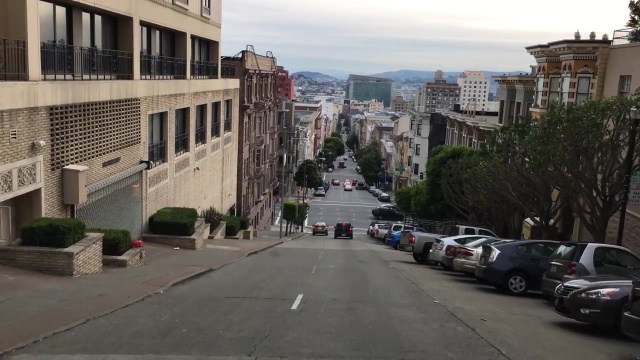} & 
\includegraphics[width=0.25\linewidth]{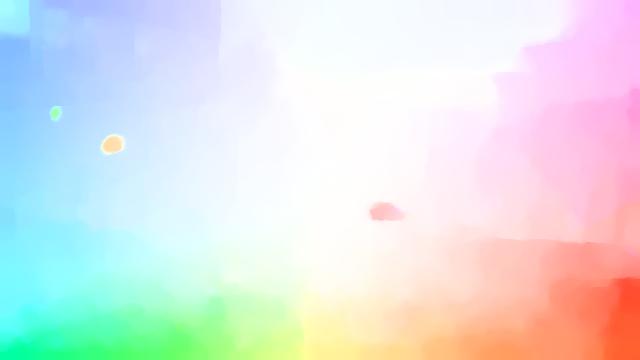} & 
\includegraphics[width=0.25\linewidth]{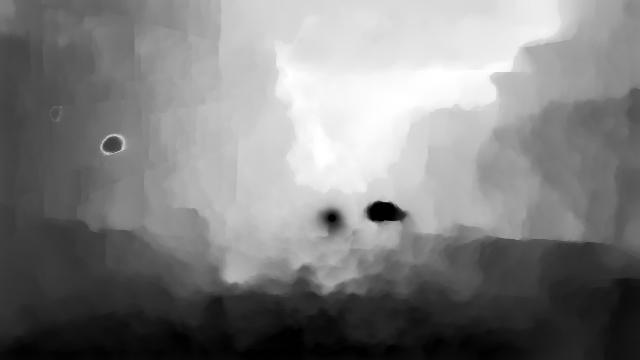} \\
\end{tabular}
\caption{Samples of image pairs and computed translational optical flow that we use to recover the relative depth. From left to right: first images, second images, translational optical flow between input two images, and relative depth of the first images.}
\label{fig:flow_depth}
\end{figure}

Note that the depth $Z$ can be estimated from either one of these equations. However, the estimate can be unstable if either $u_t$ or $v_t$ is small.  To  obtain a more robust estimate of $Z$, we square the two equations above and add them:
\begin{align}
Z=\sqrt{\frac{(-U+xW)^2 + (-V+yW)^2}{u_t^2+v_t^2}}.
\label{eq:depth_recovery}
\end{align}
Because we can only recover $(U,V,W)^T$ up to scale (see below), we can only compute the depth map of an image up to scale. To induce feature representations, we  use  depth orderings of  pixels in an image. We compute the relative depth $z\in[0, 1]$ of the pixel $P$ as its depth percentile (divided by 100) across all estimated depth values for the image. Since these percentiles are invariant to the velocity's unknown scale, we do not need to recover the absolute scale of velocity. Examples of these automatically obtained depth maps are given in Figure~\ref{fig:samples_im_depth_pairs} and Figure~\ref{fig:flow_depth}.

To compute the optical flow, we use the state-of-the-art unsupervised method~\cite{hu17robust}. It first computes sparse pixel matchings between two video frames. It then interpolates to get dense pixelwise optical flow fields from sparse matchings, where we replace the supervised edge detector~\cite{dollar15fast} with its unsupervised version~\cite{li16unsupervised}. Based on the optical flow, we use the method proposed in~\cite{bideau16its} to recover the image motion of each pixel due to translational motion only $(u_t, v_t)$, and also, the global camera motion $(U, V, W)^T$ up to an unknown scale factor. Specifically, the rotation of the camera can be estimated by finding the rotation such that the remaining motion, by removing the rotational component from the motion field, can be well-explained by angle fields, which are the angle part of the motion field. This procedure produces a translational optical flow field ($u_t,v_t$), and a set of regions in the image corresponding to the background and different object motions, along with the motion directions $(U,V,W)$ of those regions. We refer readers to~\cite{bideau16its} for more technical details.



In summary, to obtain the depth map of each frame from a video, we: 
\begin{itemize}
\item compute the optical flow ($u, v$) between a pair of frames~\cite{hu17robust};
\item estimate the translational component ($u_t, v_t$) of the optical flow and the direction of camera translation ($U, V, W$) from the optical flow, using the method of~\cite{bideau16its};
\item estimate the scene depth $Z$ using Eq.~\ref{eq:depth_recovery}, up to an unknown scale factor, from the translational component of the optical flow and the camera direction estimate and convert it to relative depth $z\in[0, 1]$.
\end{itemize}

\subsection{Predicting Relative Depth From a Single Image} 
\label{sec:network_training}

While a CNN for predicting depth from a single image is a core component of our system, we are primarily interested in relative depth prediction as a proxy task, rather than an end in itself.  We therefore select standard CNN architectures and focus on quantifying the power of the depth task for pre-training and domain adaptation, compared to using the same networks with labeled data.  Specifically, we work with variants of the standard AlexNet~\cite{krizhevsky12imagenet}, VGG16~\cite{simonyan14very}, and ResNet50~\cite{he16deep} architectures.

Given an RGB image $I$,  we need pixelwise predictions in the form of a depth image $z$, so we modify both AlexNet, VGG16, and ResNet50 to produce output  with the same spatial resolution as the input image. In particular, we consider Fully Convolutional Networks (FCNs)~\cite{shelhamer17fully} and an encoder-decoder with skip connections~\cite{godard16unsupervised,zhou17unsupervised}. Detailed discussions can be found in the experiment section.

Since the relative depth (\ie, the depth percentile) is estimated over the entire image, it is essential to feed the entire image to the CNN to make a prediction. For CityDriving and KITTI, we simply resize the input image to $224\times 416$ and $352\times 1212$. For CityScapes, we discard the bottom 20\% portion or so of each video frame containing mainly the hood of a car, which remains static over all videos and makes the relative depth estimation inaccurate (recall our relative depth estimation is mainly based on motion information). The cropped input image is then resized to $384\times 992$. During training, we employ horizontal flipping and color jittering for data augmentation.  Since relative depth serves as a proxy, rather than an end task, even though the relative depth estimation is not always correct, the network is able to tolerate some degree of noise as shown in~\cite{pathak2017move}; we can then repurpose the network's learned representation.

In all experiments, we use $L_1$ loss for each pixel when training for depth prediction, \ie, we train networks to regress the relative depth values. All AlexNet, VGG16, and ResNet50 variants are trained for 30 epochs using the Adam optimizer~\cite{Adam} with momentum of $\beta_1=0.9, \beta_2=0.999$, and weight decay of 0.0005. The learning rate is $0.0001$ and is held constant during the pre-training stage.

\newcommand{\bestres}[1]{\textbf{#1}}


\section{Experiments}
We consider three urban scene understanding tasks: semantic segmentation, joint semantic reasoning consisting of road segmentation and object detection~\cite{teichmann16multinet}, and monocular absolute depth estimation. 


\subsection{Semantic Segmentation}
We consider three datasets commonly used for evaluating semantic segmentation. Their main characteristics are summarized below:

{\bf KITTI}~\cite{ros15vision}: 100 training images, 46 testing images, spatial dimensions of 370$\times$1226, 11 classes.

{\bf CamVid}~\cite{brostow08segmentation,brostow08semantic}: 367 training images, 101 validation images, 233 testing images, spatial dimensions of $720\times 960$, 11 classes. 

{\bf CityScapes}~\cite{Cordts2016Cityscapes}: 2975 training images, 500 validation images, 1525 testing images, spatial dimensions of $1024\times 2048$, 19 classes. \new{We conduct experiments on images at half resolution.}

\begin{table}[t]
\renewcommand{\tabcolsep}{5pt}
\centering
\caption{Comparisons of mean IoU scores of AlexNet FCN32s for semantic segmentation using different self-supervised models. CS=CityScapes, K=KITTI, CV=CamVid.}
\label{tab:self_supervised_models}
\begin{tabular}{ll@{\extracolsep{15pt}}ccc}
  \toprule
	 pre-training method & supervision source & CS & K & CV \\
  \midrule
   supervised & ImageNet labels& \first{48.1} & \first{46.2} & \first{57.4} \\
   \midrule
   none & - & 40.7  & 39.6 & 44.0 \\
  \midrule
  tracking~\cite{wang2015unsupervised} & motion & 41.9 & 42.1 & 50.5\\
  moving~\cite{agrawal15learning} & ego-motion & 41.3 & 40.9 & 49.7\\
  watch-move~\cite{pathak2017move} & motion seg. & 41.5 & 40.8 & 51.7\\
  frame-order~\cite{misra2016unsupervised} & motion & 41.5 & 39.7 & 49.6 \\
  context~\cite{pathak2016context} & appearance & 39.7 & -\footnotetext{We were unable to get meaningful results of~\cite{pathak2016context} on KITTI.} & 37.8\\
  object-centric~\cite{gao16object} &appearance & 39.6 & 39.1 & 48.0\\
  colorization~\cite{larsson2016learning,larsson2017colorproxy} & appearance (color) & 42.9 & 35.8 & 53.2\\
  cross-channel~\cite{zhang2017split} & misc.& 36.8 & 40.8 & 46.3 \\
  audio~\cite{owens2016ambient} & video soundtrack& 39.6 & 40.7 & 51.5\\
  \midrule
  Ours & depth & \second{45.4} & \second{42.6} & \second{53.4}\\
  \bottomrule
\end{tabular}
\end{table}

The first two datasets are much too small to provide sufficient data for ``from scratch'' training of a deep model; CityScapes is larger, but we show below that all three datasets benefit from pre-training. We use the curated annotations of the CamVid dataset released by~\cite{kundu16feature}. As a classical CNN-based model for semantic segmentation, we report results of different variants of the Fully Convolutional Network (FCN)~\cite{shelhamer17fully}. 

We compare our results to those obtained with other self-supervision strategies surveyed in Section~\ref{sec:related}. Since only AlexNet pre-trained models are available for most of the previous self-supervised methods, we also train an AlexNet. During training, the inputs are random crops of $352\times 352$ for KITTI and $704\times 704$ for CamVid. Each FCN32s using different pre-training models is trained for 600 epochs with a batch size of 16 using 4 GPUs. For CityScapes, the inputs to the network are random crops of $512\times 512$. Each FCN32s is trained for 400 epochs with a batch size of 16. In addition to the random crops, random horizontal flips and color jittering are also performed. The CNNs at this stage (learning segmentation) are trained or fine-tuned using the Adam optimizer, where weight decay is 0.0005. For the learning rate, we use 0.0001 and decrease it by factor of 10 at the 400th epoch (300th epoch for CityScapes). 

Quantitative comparisons can be found in Table~\ref{tab:self_supervised_models}\footnote{We were unable to get meaningful results with~\cite{pathak2016context} on KITTI and with~\cite{jayaraman15learning} on all three segmentation datasets.}. Our pre-trained model performs significantly better than the model learned from scratch on all three datasets, validating the effectiveness of our pre-training. Moreover, we obtain new state-of-the-art results on all three urban scene segmentation datasets among methods that use self-supervised pre-training. In particular, our model outperforms all other self-supervised models with motion cues (the first four self-supervised models in Table~\ref{tab:self_supervised_models}). 

\subsection{Ablation Studies}
We perform ablation studies using VGG16 FCN32s on the semantic segmentation datasets. Specifically, we study the following aspects.

\begin{figure}[t]
\centering
\begin{tabular}{@{\extracolsep{0pt}}cc}
\includegraphics[width=0.43\linewidth]{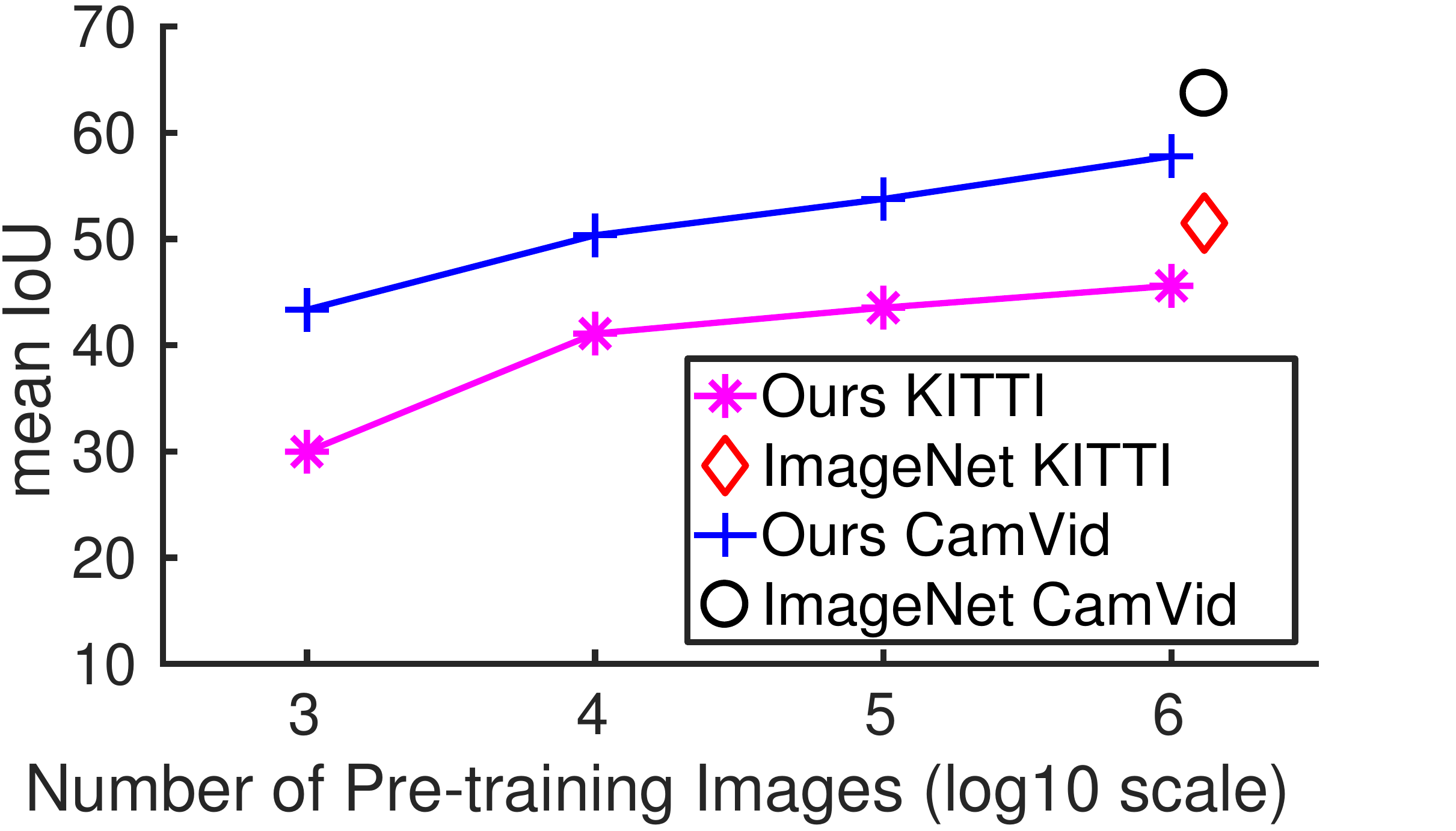} &
\includegraphics[width=0.43\linewidth]{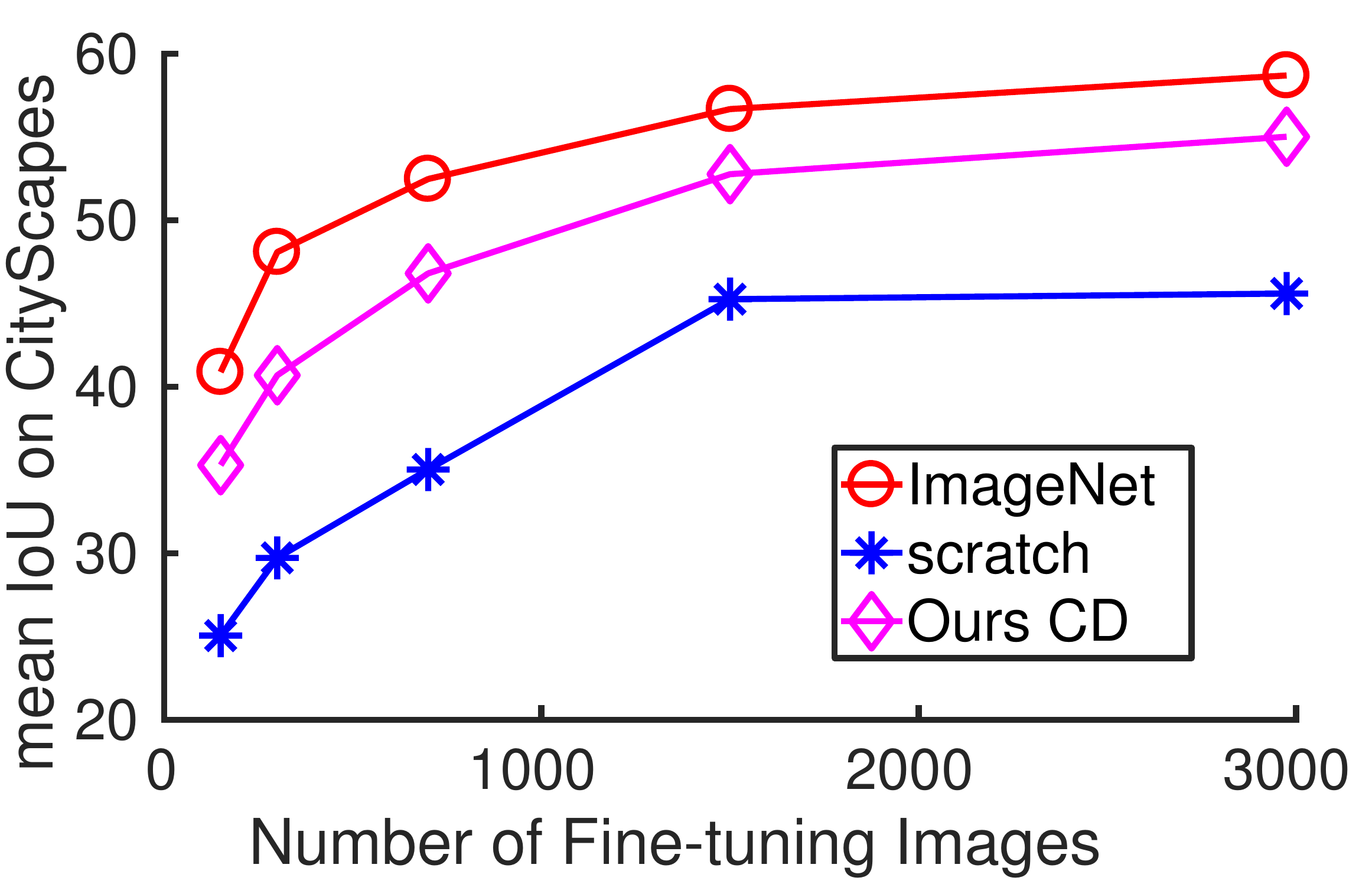} \\
(a) & (b) \\
\end{tabular}
\caption{Ablation studies of performance by (a) varying number of pre-training images on KITTI and CamVid, (b) varying number of fine-tuning images of CityScapes.}
\label{fig:ablation_study}
\end{figure}

\noindent\textbf{Number of pre-training images.} 
Figure~\ref{fig:ablation_study}(a) demonstrates that the performance of our depth pre-trained model scales linearly with the log of the number of pre-training images on CamVid, which is similar to the conclusion of~\cite{pathak2017move}. 

On KITTI, our pre-trained model initially has a big performance boost when the number of pre-training images increases from 1K to 10K. With enough data (more than 10K), the performance also scales linearly with the log of the number of the pre-training images.

\noindent\textbf{Number of fine-tuning images.} Figure~\ref{fig:ablation_study}(b) shows that every model (ImageNet, scratch, our depth pre-trained model) benefits from more fine-tuning data on the CityScapes dataset. 
For both ImageNet and our depth pre-trained models, it suggests that more fine-tuning data is also beneficial for transferring the previously learned representations to a new task.


\subsection{Domain Adaptation by Pre-Training}

\newcommand{\cityscapes}{\small{CS}}
\newcommand{\camvid}{\small{CV}}
\newcommand{\kitti}{\small{K}}
\begin{table*}[t]
\renewcommand{\tabcolsep}{4pt}
\centering
\caption{Mean IoU scores of semantic segmentation using different architectures on different datasets. CD=CityDriving, CS=CityScapes, CV=CamVid, and K=KITTI.}
\label{tab:semantic_seg}
\begin{tabular}{@{\extracolsep{6pt}}cccccccccc}
\toprule
\multirow{2}{*}{pre-training} & \multicolumn{3}{c}{FCN32s} & \multicolumn{3}{c}{FCN16s} & \multicolumn{3}{c}{FCN8s} \\
\cline{2-4}\cline{5-7}\cline{8-10}
& \cityscapes & \camvid & \kitti & \cityscapes & \camvid & \kitti & \cityscapes & \camvid & \kitti \\
\midrule
ImageNet & \first{58.7} & \first{63.7} & \first{51.5} & \first{62.9} & \first{65.9} & \first{55.3} & \first{63.4} & \first{67.0} & \first{56.4} \\
scratch & 45.4 & 41.0 & 32.4 & 51.3 & 44.1 & 33.1 & 51.6 & 44.3 & 34.2 \\
\midrule
Ours CD & 55.0 & \third{57.8} & 45.6 & \third{57.6} & \second{59.0} & 47.7 & \third{59.8} & \second{60.3} & 48.6 \\
Ours CD+K & \third{56.0} & \second{58.5} & \third{46.0} & 56.9 & \third{58.8} & \second{48.2} & 58.9 & \third{60.1} & \third{49.0} \\
Ours CD+CS & \second{56.2} & \second{58.5} & \second{47.4} & \second{58.5} & \third{58.8} & \third{47.8} & \second{60.5} & 59.9 & \second{49.6} \\
\bottomrule
\end{tabular}
\end{table*}

\begin{figure}[h]
\renewcommand{\arraystretch}{0.6}
\renewcommand{\tabcolsep}{.05mm}
\centering
\begin{tabular}{@{\extracolsep{0.6pt}}ccccc}
	    \includegraphics[height=0.12\linewidth]{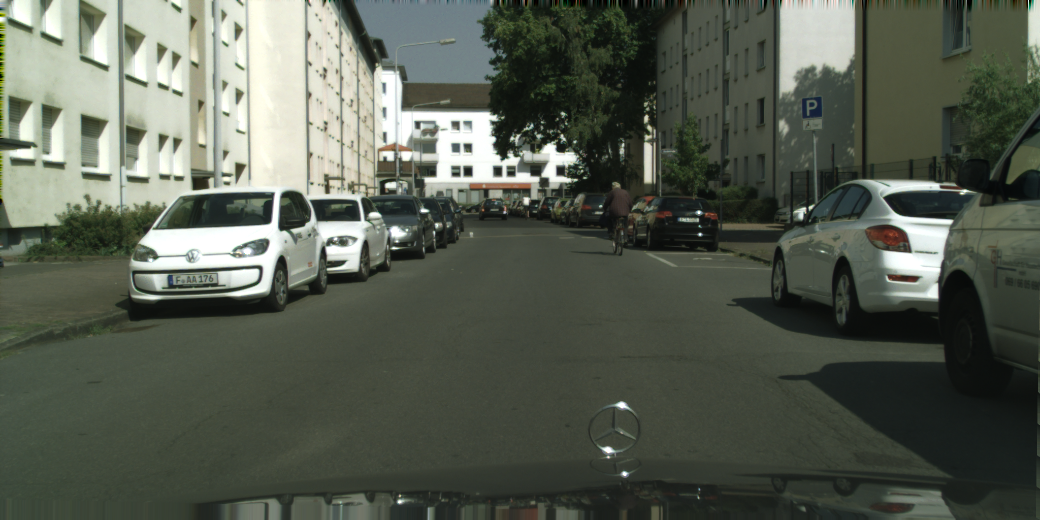} & 
    \includegraphics[height=0.12\linewidth]{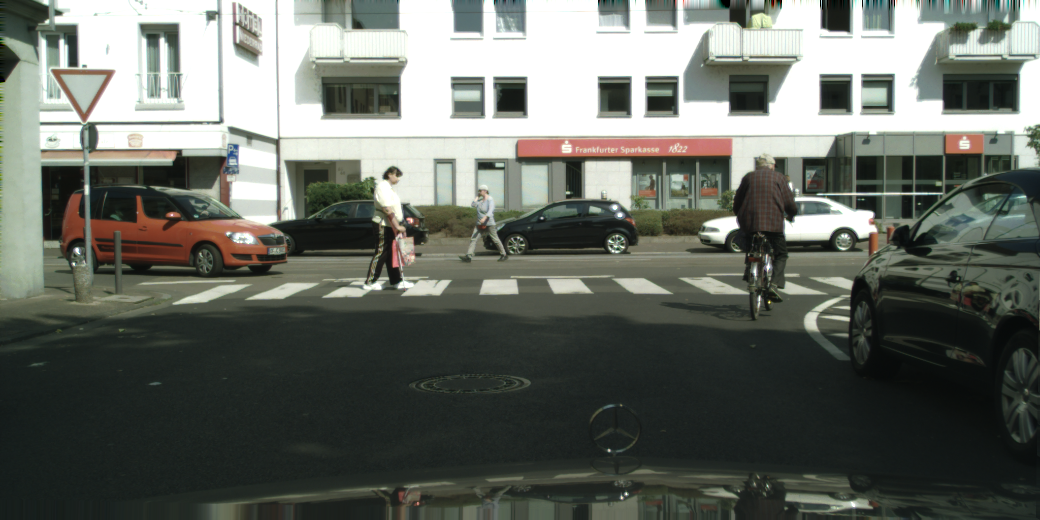} & 
    \includegraphics[height=0.12\linewidth]{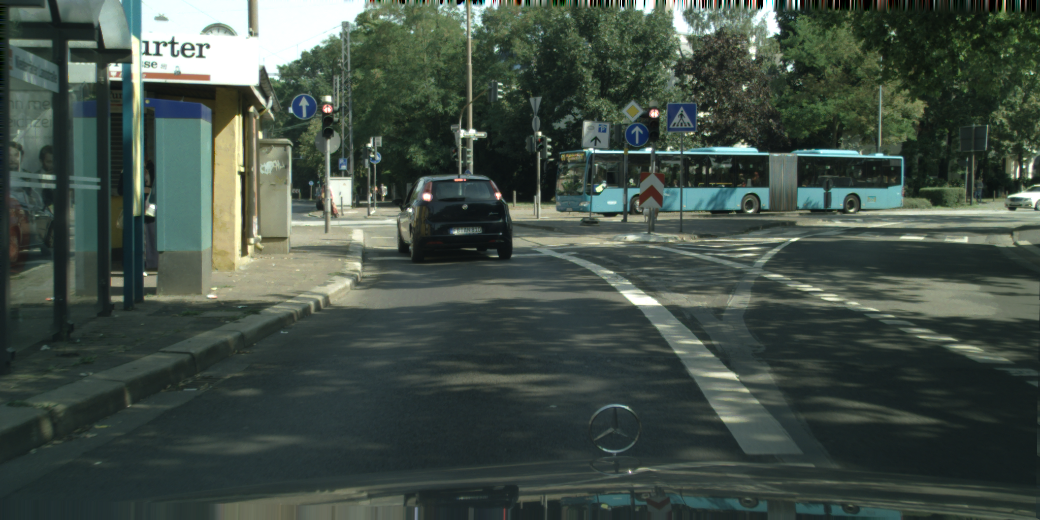} & 
    \includegraphics[height=0.12\linewidth]{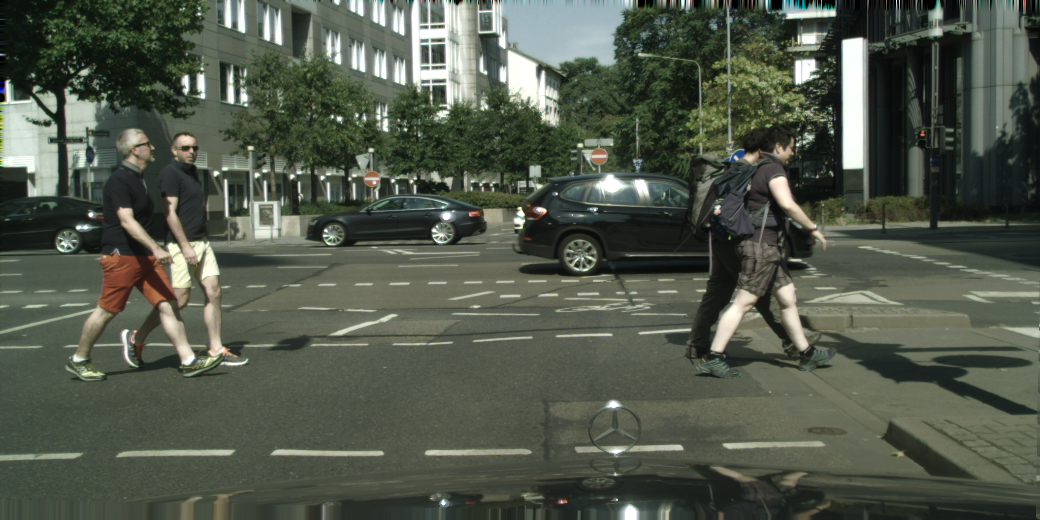} \\
    \includegraphics[height=0.12\linewidth]{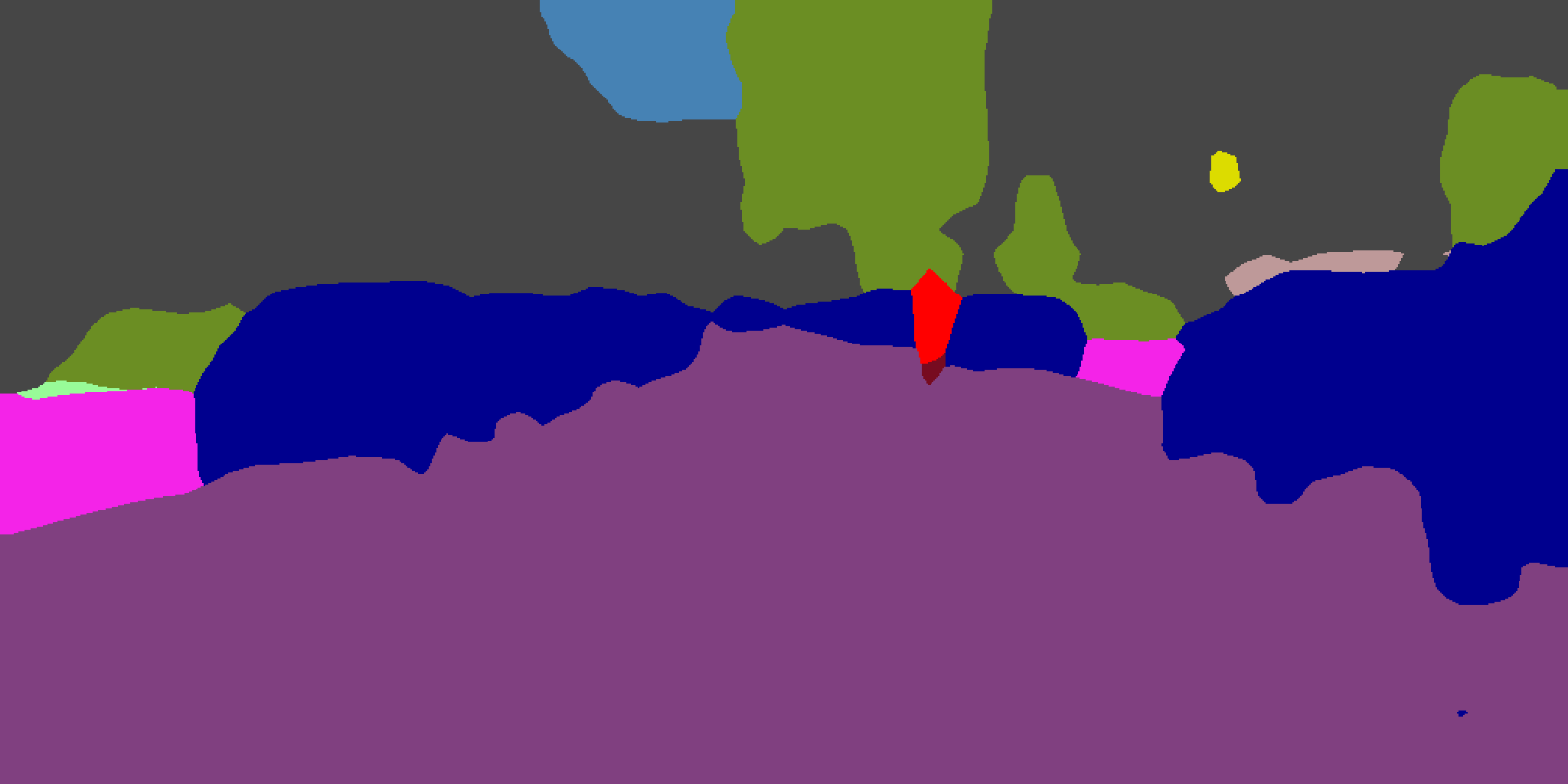} & 
    \includegraphics[height=0.12\linewidth]{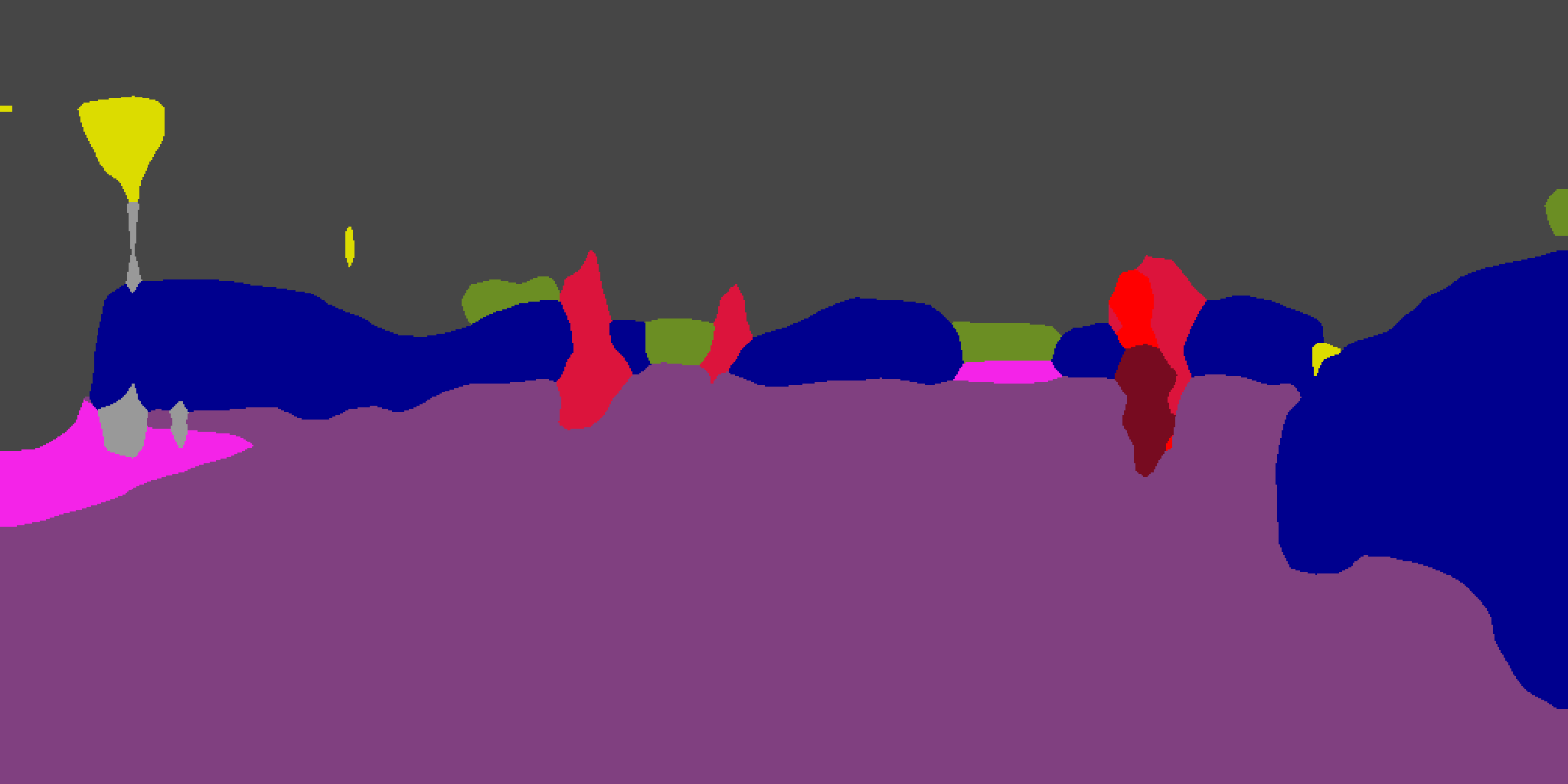} & 
    \includegraphics[height=0.12\linewidth]{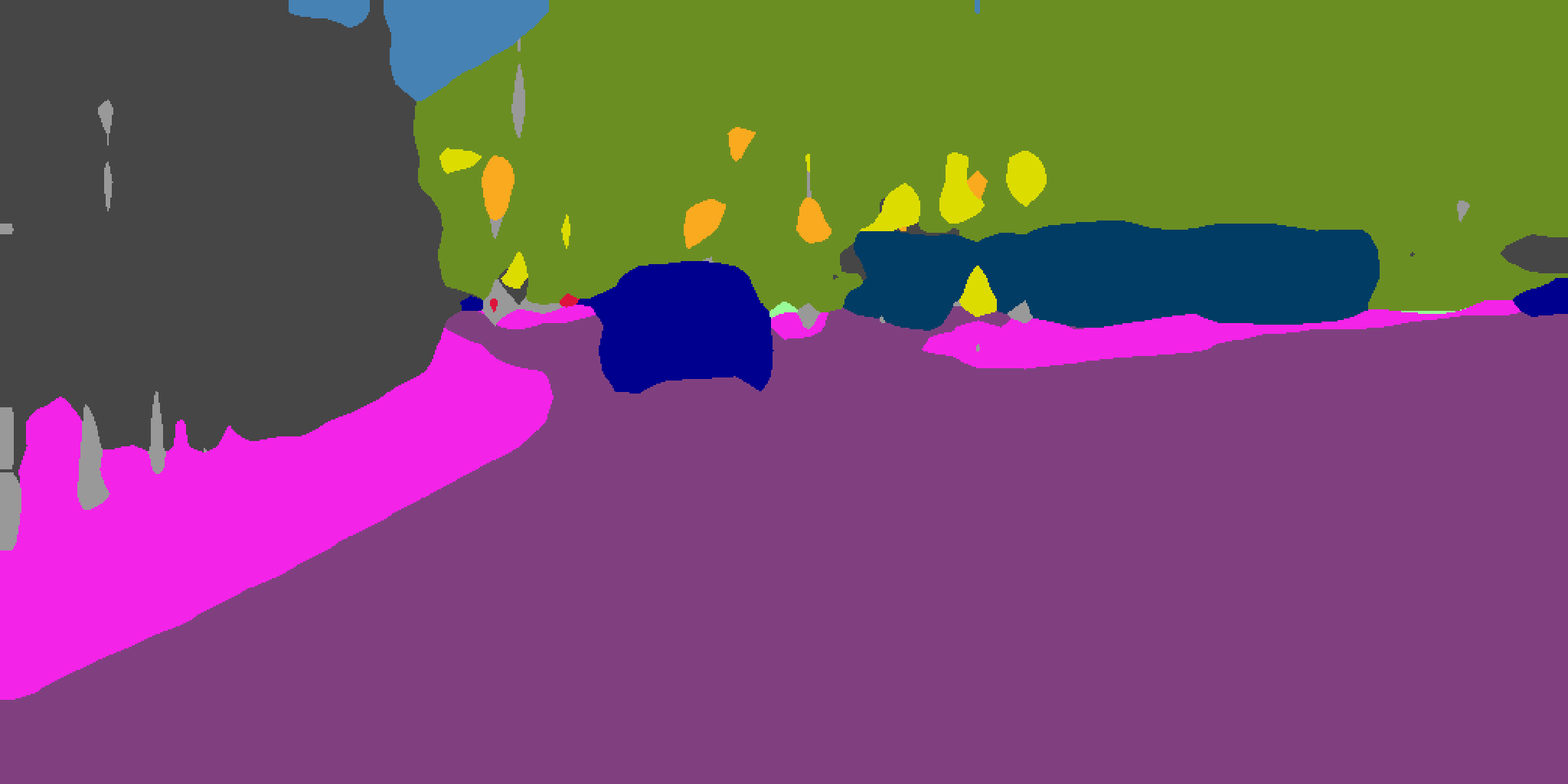} & 
    \includegraphics[height=0.12\linewidth]{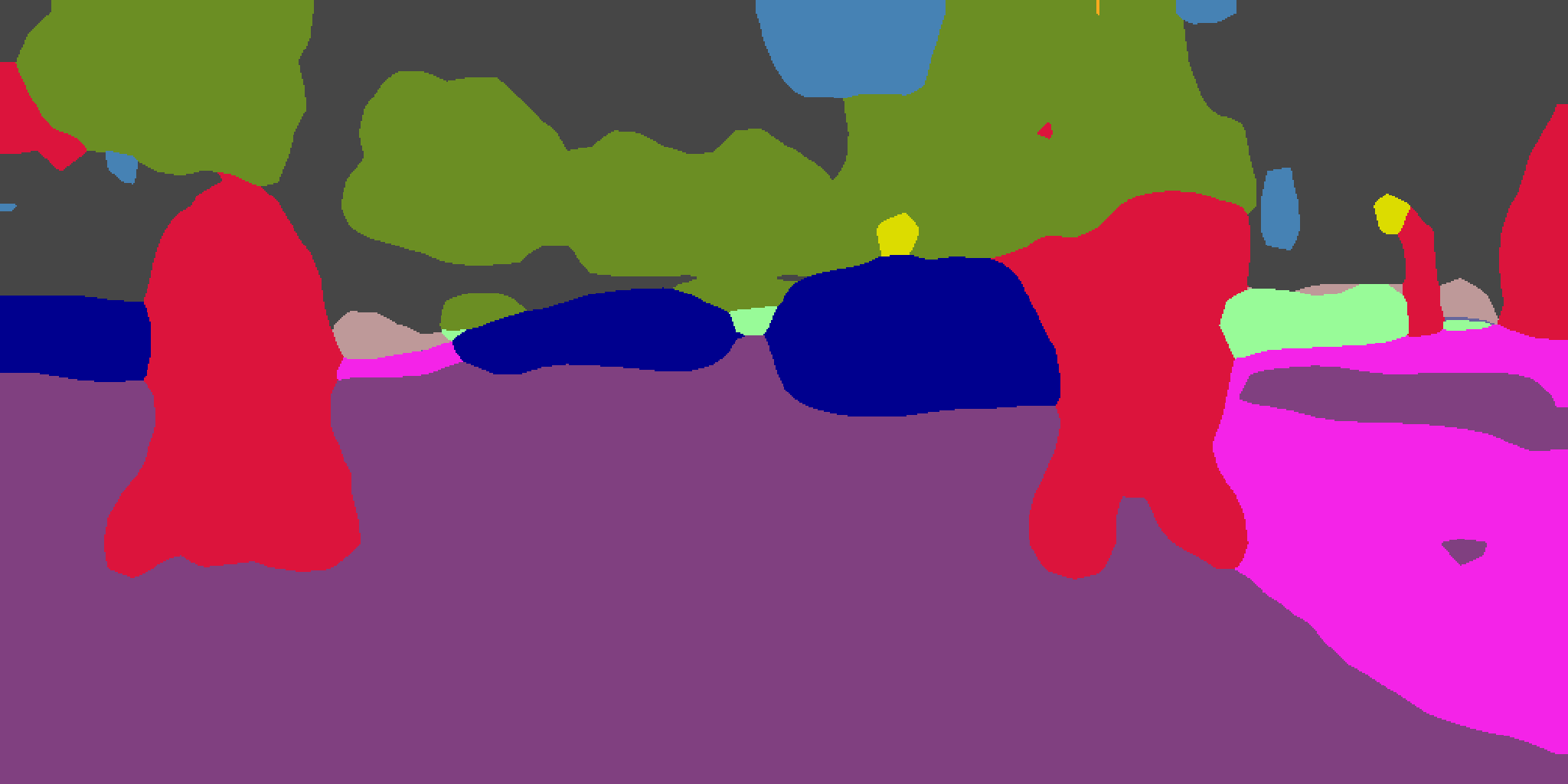} \\
    \includegraphics[height=0.12\linewidth]{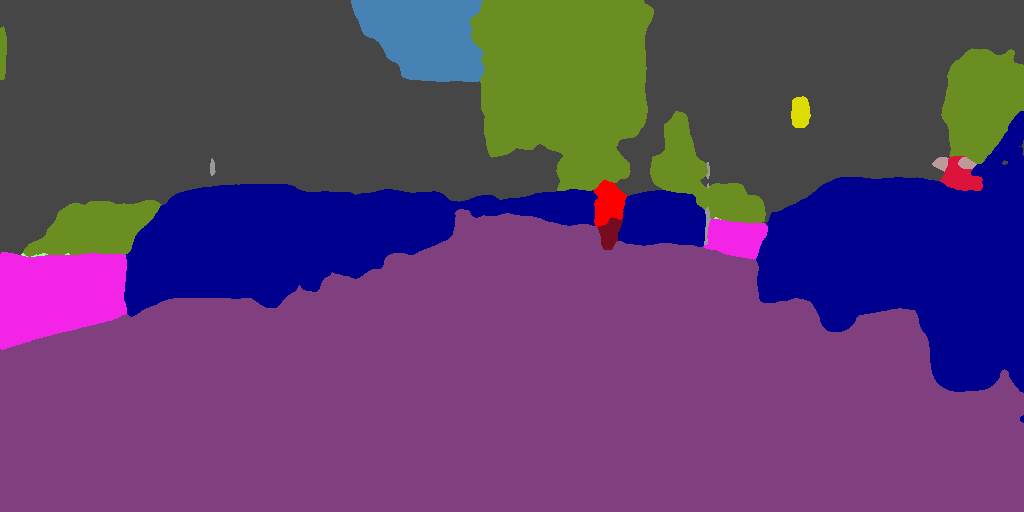} & 
    \includegraphics[height=0.12\linewidth]{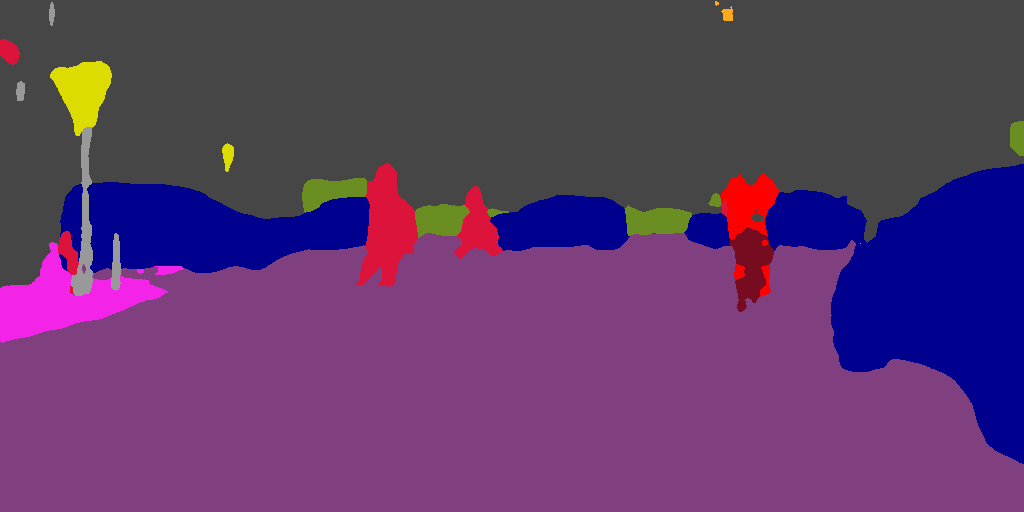} & 
    \includegraphics[height=0.12\linewidth]{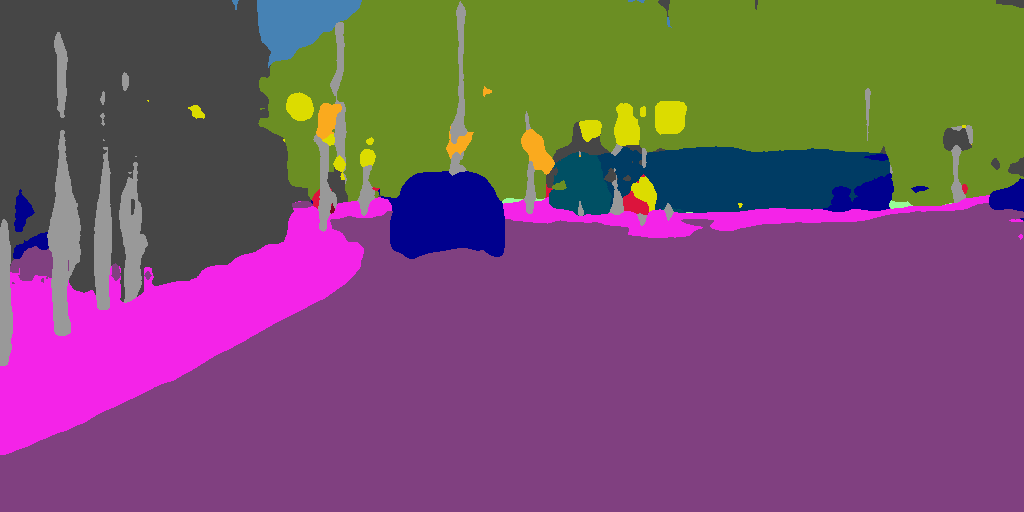} & 
    \includegraphics[height=0.12\linewidth]{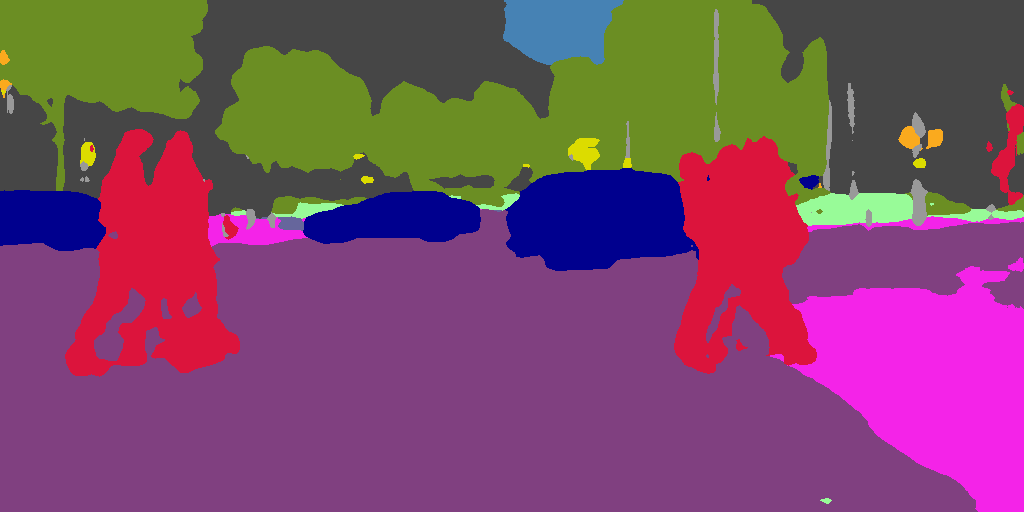} \\
    \includegraphics[height=0.12\linewidth]{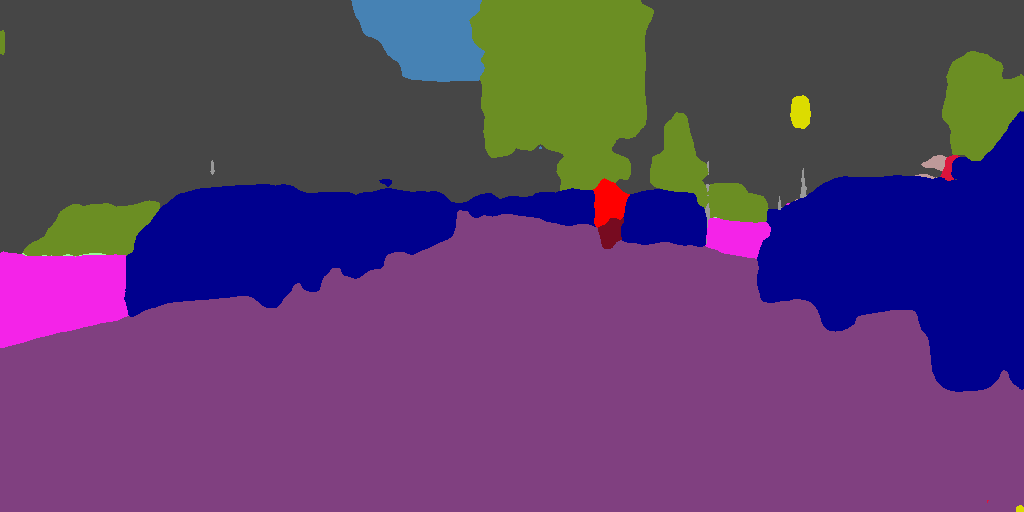} & 
    \includegraphics[height=0.12\linewidth]{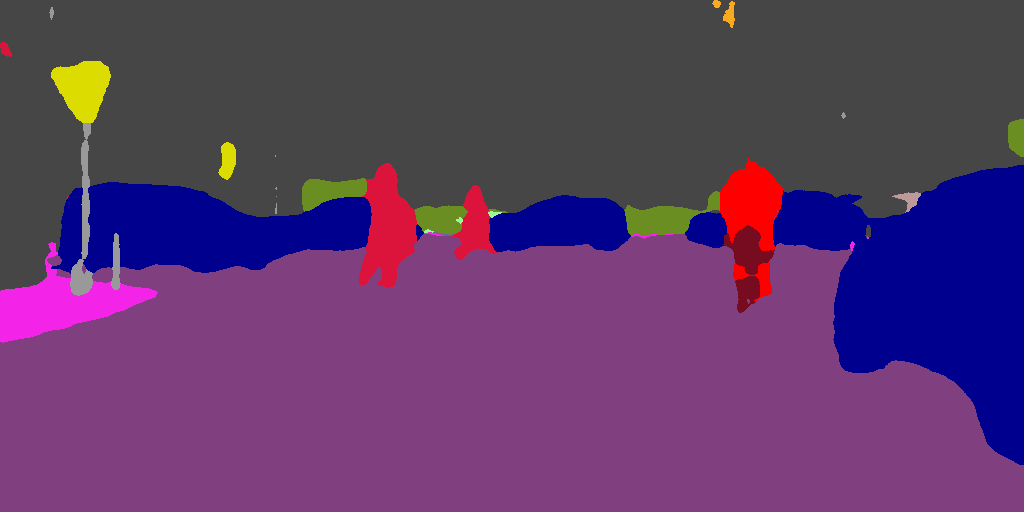} & 
    \includegraphics[height=0.12\linewidth]{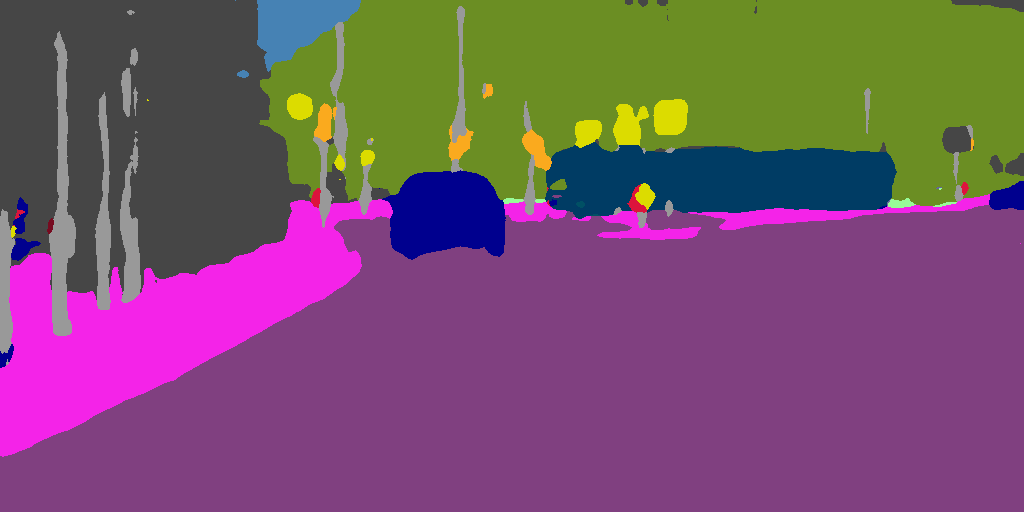} & 
    \includegraphics[height=0.12\linewidth]{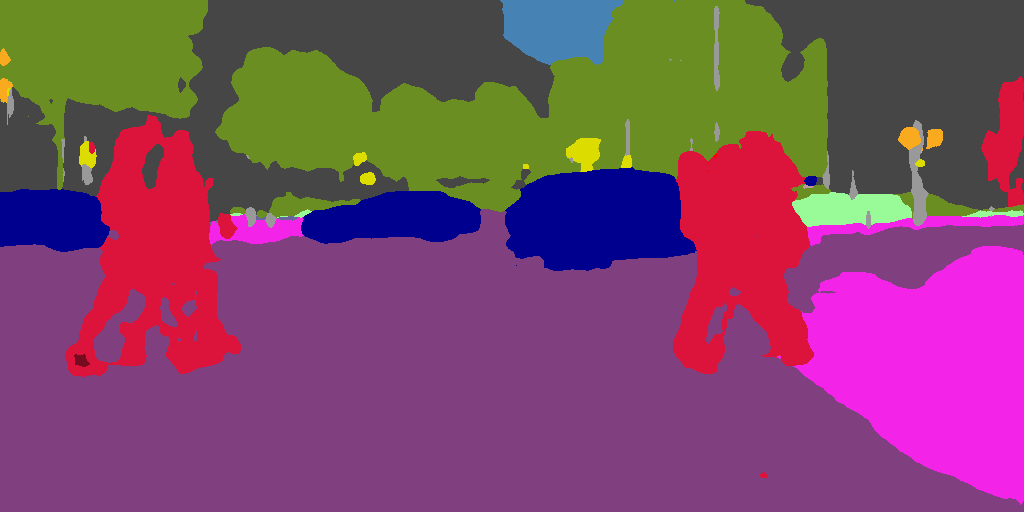} \\
    \includegraphics[height=0.12\linewidth]{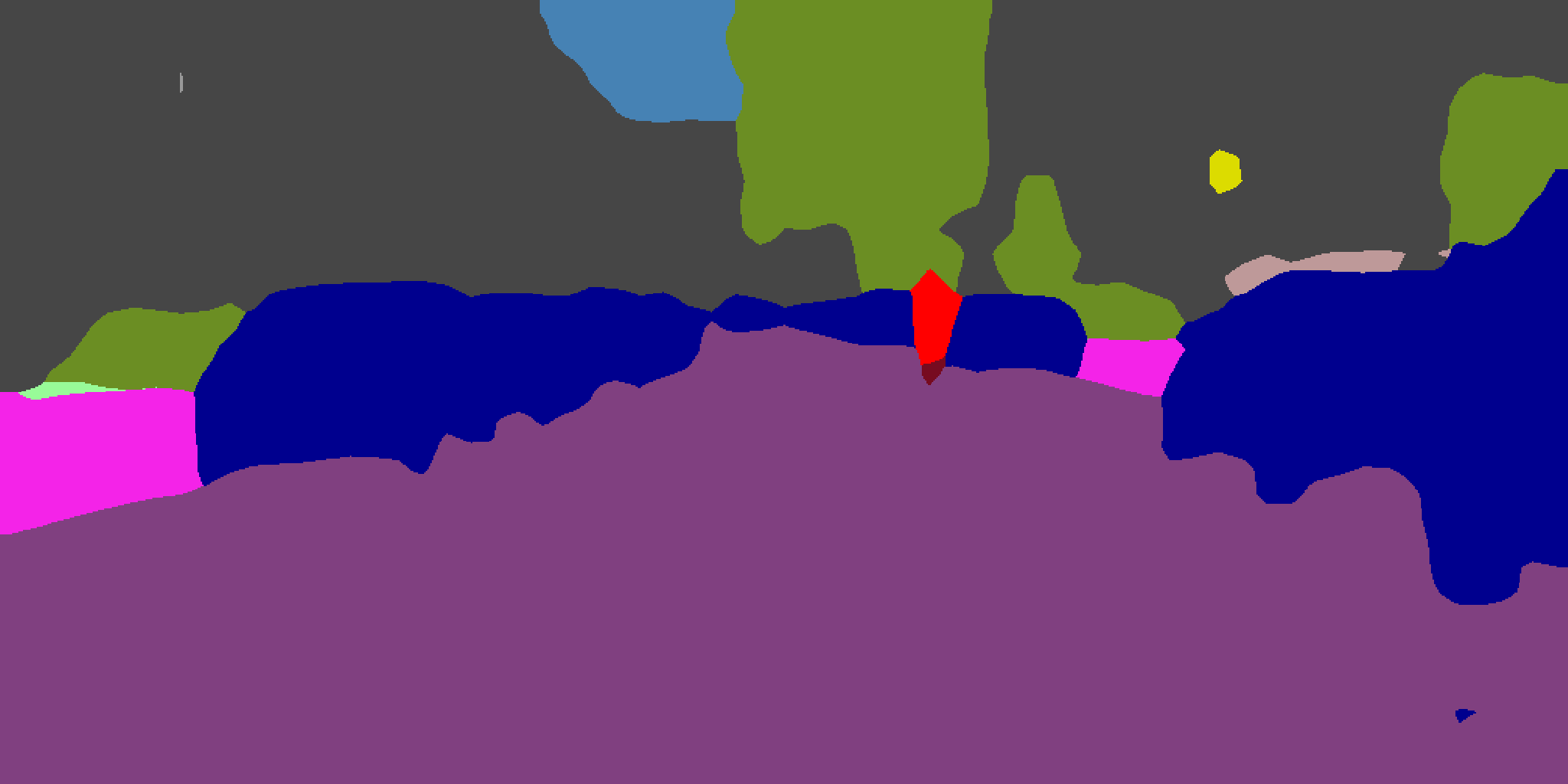} & 
    \includegraphics[height=0.12\linewidth]{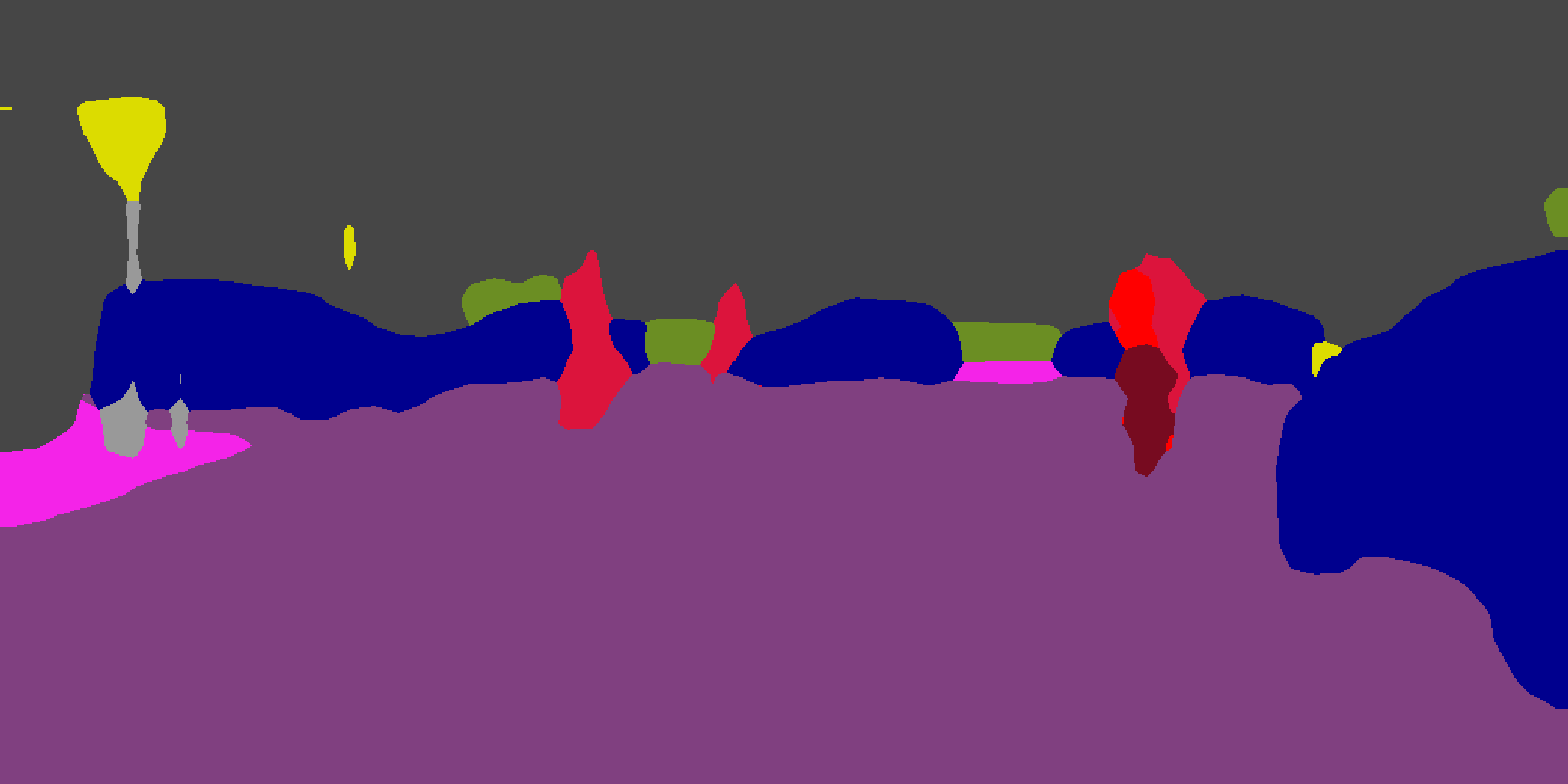} & 
    \includegraphics[height=0.12\linewidth]{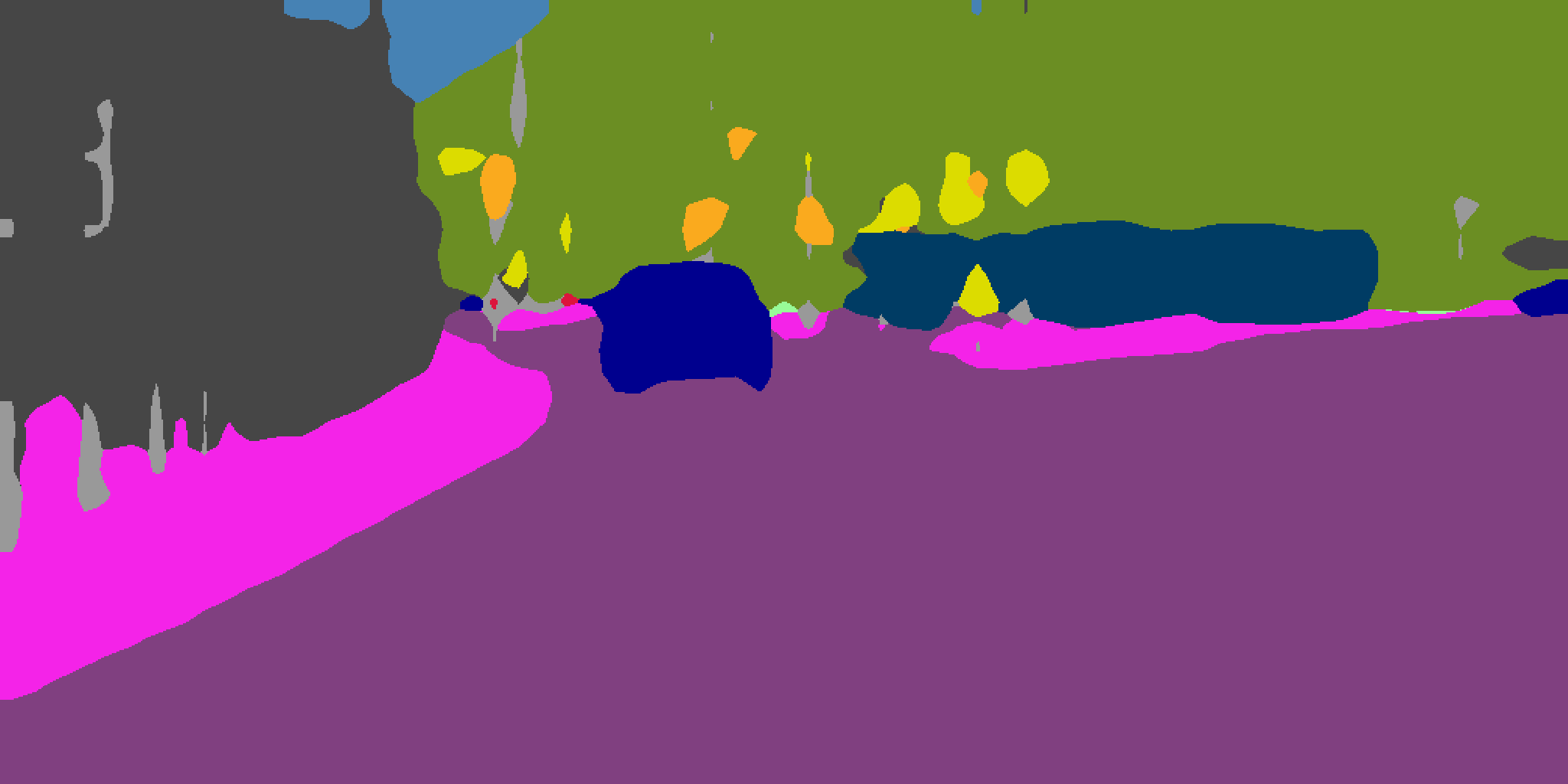} & 
    \includegraphics[height=0.12\linewidth]{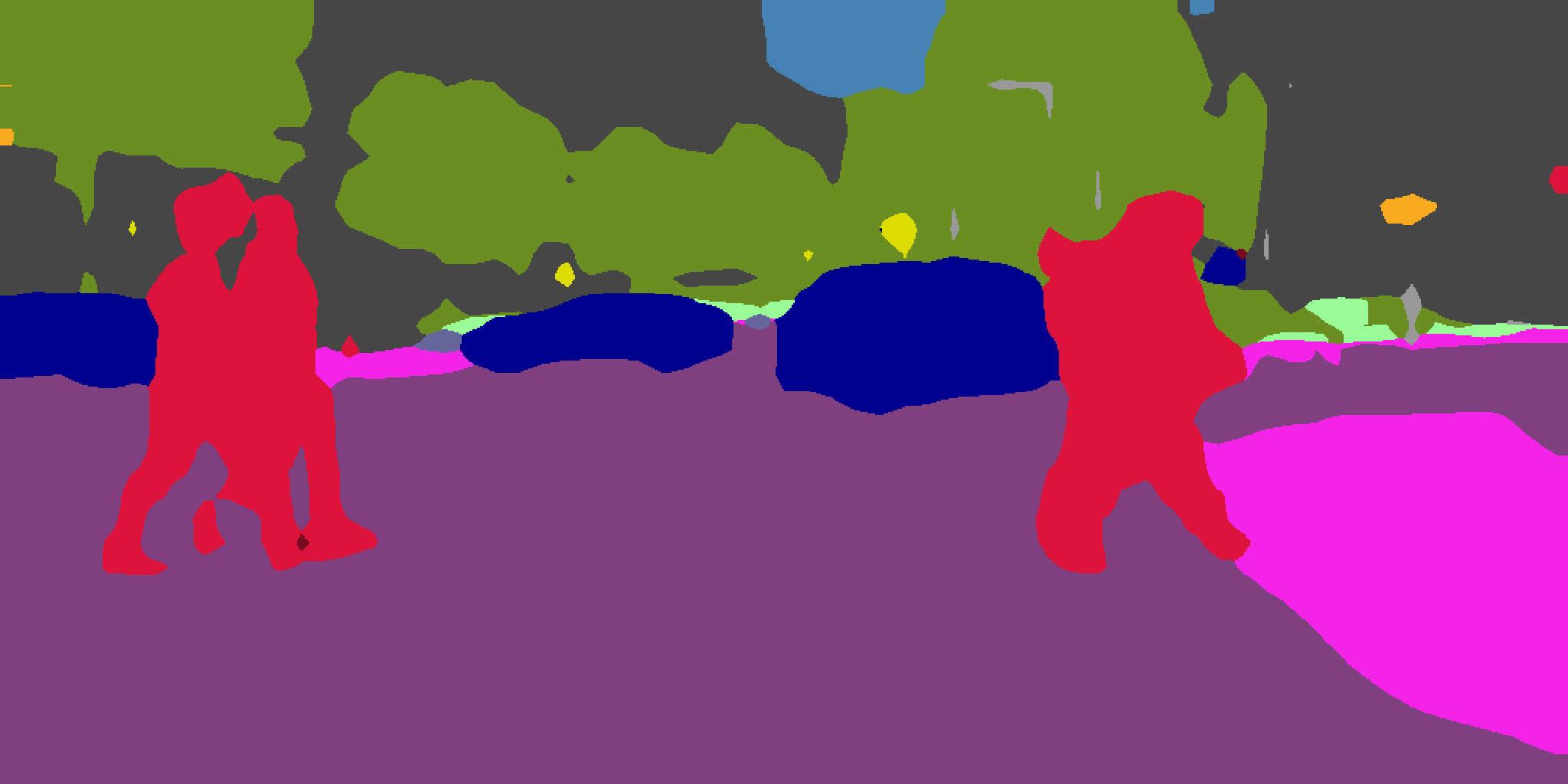} \\
	\includegraphics[height=0.12\linewidth]{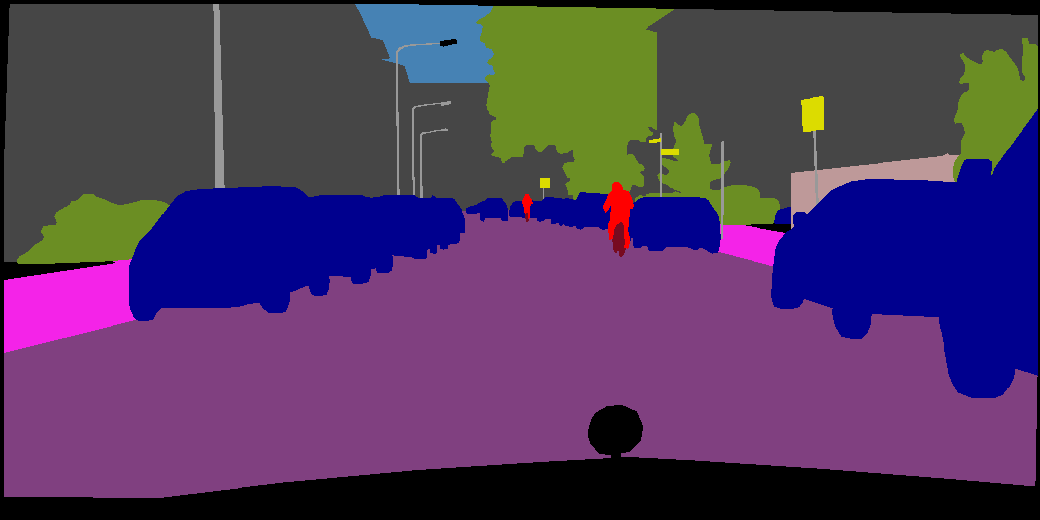} & 
    \includegraphics[height=0.12\linewidth]{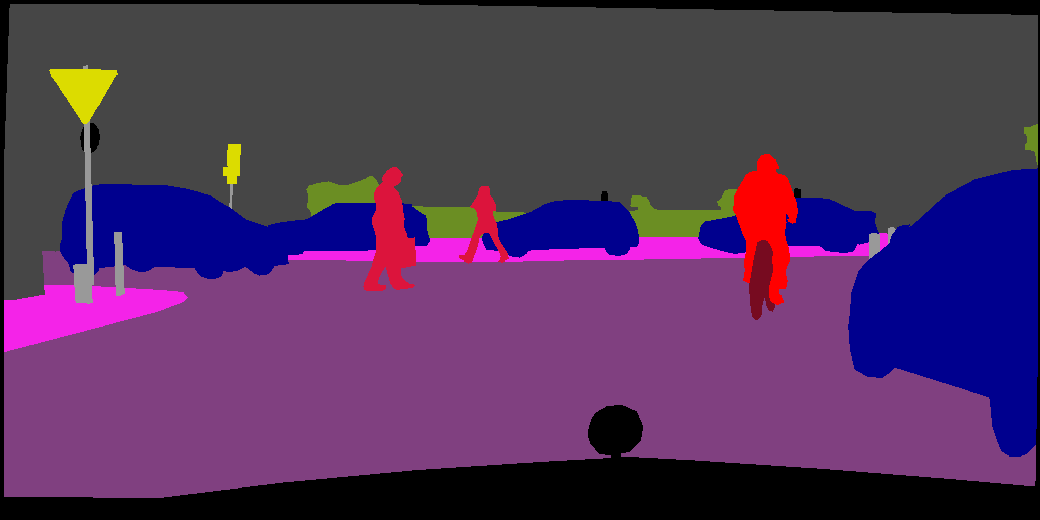} & 
    \includegraphics[height=0.12\linewidth]{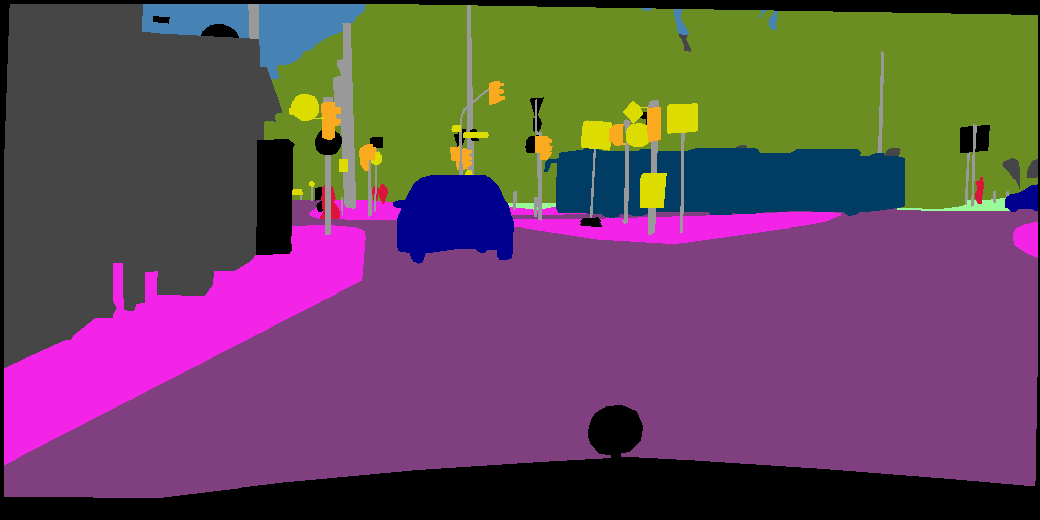} & 
    \includegraphics[height=0.12\linewidth]{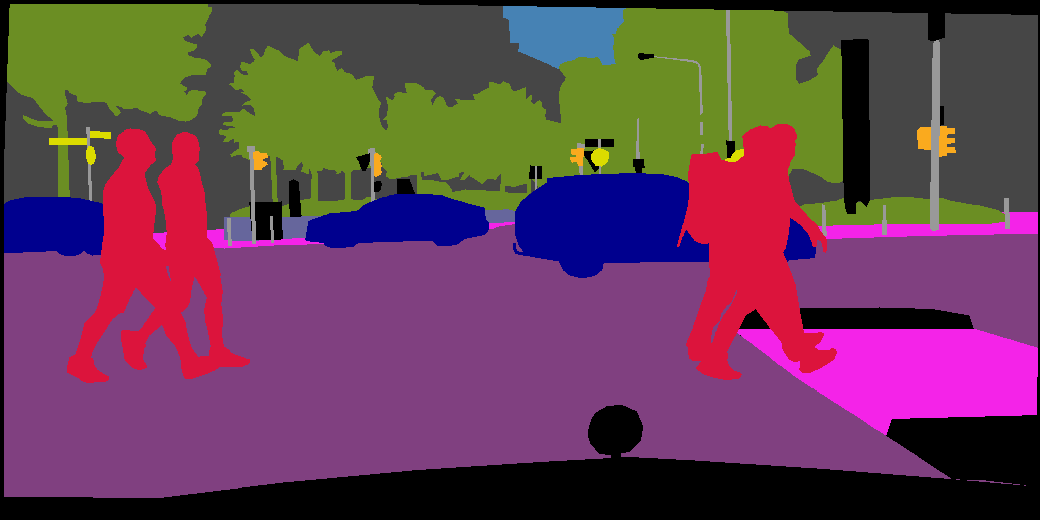} \\
\end{tabular}
\caption{Qualitative semantic segmentation results on CityScapes. From top to bottom: input images, predictions of FCN8s with no pre-training, our FCN8s pre-trained on CityDriving, our FCN8s pre-trained on CityDriving adapted to CityScapes,  ImageNet FCN8s, and ground-truth annotations. The difference between the 2nd and 3rd rows shows a clear benefit of pre-training with relative depth prediction. The difference between 3rd rows and 4th rows shows the benefit our unsupervised domain adaptation using pre-training (\eg, the bicyclist and the bus in the second and third columns, respectively).}
\label{fig:cityscapes_res}
\end{figure}

In the experiments described above, the two stages (pre-training on self-supervised depth prediction, followed by supervised training for segmentation) rely on data that come from significantly different domains. The self-supervised learning uses videos obtained from moving through North American cities. In contrast, none of the target dataset images were collected in the same geographic locations. For instance, CityScapes includes data from driving in German cities. Thus, in addition to a shift in task, the fine-tuning of the network for segmentation must also deal with a \emph{domain shift} in the input. 


CityScapes~\cite{Cordts2016Cityscapes} and KITTI~\cite{Geiger2012CVPR} make available video sequences that give temporal context to every image in the dataset. None of these extra frames are labeled, but we can leverage them in the following way. Before training the network on segmentation, we fine-tune it, using  the same self-supervised relative depth prediction task described in Section~\ref{sec:self_supervised_depth}, on these videos. Our intuition is that this may induce some of the modifications in the network that reflect the changing distribution of the input. Then, we proceed as before to train the fine-tuned representation on the semantic segmentation data.
Specifically, we fine-tune different FCNs variants based on VGG16 sequentially, \ie, from FCN32s to FCN16s, and finally to FCN8s. For FCN32s, the training procedure is identical to AlexNet FCN32s described earlier. FCN16s and FCN8s are trained for the same number of epochs as FCN32s, where the learning rate is set to 0.00002 and 0.00001, respectively, and kept constant during training.

The effectiveness of our unsupervised domain adaptation for semantic segmentation can be found in Table~\ref{tab:semantic_seg}. The last two rows of demonstrate that such fine-tuning can consistently improve the performance of a self-supervised model over all FCN variants on both CityScapes and KITTI, validating its effectiveness as a domain adaptation approach. Interestingly, we can see that while fine-tuning is helpful for FCN32s on CamVid initially, it does not help much for FCN16s and FCN8s. Perhaps this is due to the domain gap between CamVid and CityScapes/KITTI. Qualitative semantic segmentation results can be found in Fig.~\ref{fig:cityscapes_res}.

\subsection{Joint Semantic Reasoning}

Joint semantic reasoning is important for urban scene understanding, especially with respect to tasks such as autonomous driving~\cite{teichmann16multinet}. We investigate the effectiveness of our pre-trained model and the unsupervised strategy of domain adaptation using the MultiNet architecture~\cite{teichmann16multinet} for joint road segmentation and car detection.\footnote{We use the author's released code~\url{https://github.com/MarvinTeichmann/MultiNet}. As the scene classification data is not publicly available, we only study road segmentation and car detection here.} MultiNet consists of a single encoder, using the VGG16 as backbone, and two sibling decoders for each task. For road segmentation, the decoder contains three upsampling layers, forming an FCN8s. The car detection decoder directly regresses the coordinates of objects. 
Following~\cite{teichmann16multinet}, the entire network is jointly trained using the Adam optimizer, using a learning rate of 0.00005 and weight decay of 0.0005 for 200K steps. We refer readers to~\cite{teichmann16multinet} for more technical details.

\begin{table}[t]
\renewcommand{\tabcolsep}{5pt}
\centering
\caption{Results of joint semantic reasoning, including road segmentation and car detection.}
\label{tab:joint_semantic_reasoning}
\begin{tabular}{@{\extracolsep{5pt}}cccccc}
\toprule
\multirow{2}{*}{pre-training} & \multicolumn{2}{c}{Road Segmentation} & \multicolumn{3}{c}{Car Detection (AP)} \\
\cline{2-3}\cline{4-6}
 & $F_1$ & AP & Easy & Medium & Hard \\
\midrule
ImageNet & \first{96.33} & \first{92.26} & \first{95.59} & \first{86.43} & \first{72.28} \\
scratch & 93.78 & 91.37 & 89.37 & 79.93 & 66.02 \\
\midrule
Ours CD & 94.74 & 92.13 & 92.84 & 84.73 & 69.47 \\
Ours CD+K & \second{95.66} & \second{92.14} & \second{94.31} & \second{85.72} & \second{70.50} \\
\bottomrule
\end{tabular}
\end{table}

We replace the ImageNet-trained VGG16 network with a randomly initialized one and our own VGG16 pre-trained on CityDriving using relative depth. For the road segmentation task, there are 241 training and 48 validation images. For car detection, there are 7K training images 481 validation images.  Detailed comparisons on the validation set can be found in Table~\ref{tab:joint_semantic_reasoning}. We use the $F_1$ measure and Average Precision (AP) scores for road segmentation evaluation and AP scores for car detection. AP scores for different car categories are reported separately. We can clearly see that our pre-trained model (Ours CD) consistently outperforms the randomly initialized model (scratch in Table~\ref{tab:joint_semantic_reasoning}). Furthermore, by using the domain adaptation strategy via fine-tuning on the KITTI raw videos (Ours CD+K), we can further close the gap between an ImageNet pre-trained model. Remarkably, after fine-tuning, the $F_1$ score of road segmentation and AP scores for easy and medium categories of our pre-trained model are pretty close to the ImageNet counterpart's. (See last row of Table~\ref{tab:joint_semantic_reasoning}.)

\subsection{Monocular Absolute Depth Estimation}

For the monocular absolute depth estimation, we adopt the U-Net architecture~\cite{Ronneberger15UNet} similar to~\cite{godard16unsupervised,zhou17unsupervised}, which consists of a fully convolutional encoder and another fully convolutional decoder with skip connections. In order to use an ImageNet pre-trained model, we replace the encoder with the VGG16 and ResNet50 architectures. We use the training and validation set of~\cite{godard16unsupervised}, containing 22.6K and 888 images, respectively. We evaluate our model on the Eigen split~\cite{eigen14depth,godard16unsupervised}, consisting of 697 images, where ground-truth absolute depth values are captured using LiDAR at sparse pixels. Unlike~\cite{godard16unsupervised}, which uses stereo image pairs as supervision to train the network, or~\cite{zhou17unsupervised}, which uses neighboring video frames as supervision to train the network (\emph{yet camera intrinsic parameters are required}), we use the absolute sparse LiDAR depth values to fine-tune our network. The entire network (either VGG16 or ResNet50 version) is trained for 300 epochs using the Adam optimizer with a weight decay of 0.0005. The initial learning rate is 0.0001 and decreased by factor of 10 at the 200th epoch.

\begin{table*}[t]
\renewcommand{\tabcolsep}{0pt}
\centering
\small
\caption{Monocular depth estimation on the KITTI dataset using the split of Eigen~\etal~\cite{eigen14depth} (range of 0-80m). \new{For model details, Arch.=Architecture, A=AlexNet, V=VGG16, and R=ResNet50. For training data, Class.=classification, I=ImageNet, CD=CityDriving, K=KITTI, CS=CityScapes. \textbf{pp} indicates test-time augmentation by horizontally flipping the input image.}}
\label{tab:mono_depth_kitti2015_eigen}
\begin{tabular}{c@{\extracolsep{3pt}}ccccc@{\extracolsep{5.5pt}}cccccccc@{}}
\toprule
\multirow{2}{*}{Method} & \multirow{2}{*}{Arch.} & \multicolumn{4}{c}{Training Data} & \multicolumn{4}{c}{Error Metrics} & \multicolumn{3}{c}{Accuracy Metrics} \\
\cline{3-6}\cline{7-10}\cline{11-13}
 & & \scriptsize{Class.} & \scriptsize{Stereo} & \scriptsize{Video} & \scriptsize{GT} & \scriptsize{Abs Rel} & \scriptsize{Sq Rel} & \scriptsize{RMSE} & \scriptsize{RMSE log} & \scriptsize{$\delta<1.25$} & \scriptsize{$\delta<1.25^2$} & \scriptsize{$\delta<1.25^3$} \\
\midrule
\cite{eigen14depth} & A & I & - & - & K & 0.203 & 1.548 & 6.307 & 0.282 & 0.702 & 0.890 & 0.958 \\
\cite{liu16learning} & A & I & - & - & K & 0.202 & 1.614 & 6.523 & 0.275 & 0.678 & 0.895 & 0.965 \\
\cite{godard16unsupervised}+\textbf{pp} & V & - & CS+K & - & - & 0.118 & 0.923 & 5.015 & 0.210 & 0.854 & 0.947 & 0.976 \\
\cite{godard16unsupervised}+\textbf{pp} & R & - & CS+K & - & - & 0.114 & 0.898 & 4.935 & 0.206 & 0.861 & 0.949 & 0.976 \\
\cite{zhou17unsupervised} & V & - & - & CS+K & - & 0.198 & 1.836 & 6.565 & 0.275 & 0.718 & 0.901 & 0.960 \\
\cite{kuznietsov17semi} & R & I & K & - & K & \first{0.113}	& \first{0.741} & \first{4.621} & \first{0.189} & \first{0.862}	& \first{0.960} & \first{0.986} \\
\midrule
Ours & V & I & - & - & K & 0.157 & 1.115 & 5.546 & 0.233 & 0.768 & 0.922 & 0.974 \\
Ours & V & - & - & - & K & 0.163 & 1.241 & 5.649 & 0.238 & 0.765 & 0.918 & 0.970 \\
Ours & V & - & - & CD & K & 0.154 & 1.117 & 5.499 & 0.228 & 0.775 & 0.928 & 0.976 \\
Ours & V & - & - & CD+K & K & \textbf{0.148} & \textbf{1.056} & \textbf{5.317} & \textbf{0.221} & \textbf{0.791} & \textbf{0.932} & \textbf{0.977} \\
\midrule
Ours & R & I & - & - & K & 0.128 & 0.933 & 5.073 & 0.203 & 0.827 & 0.945 & 0.980 \\
Ours & R & - & - & - & K & 0.131 & 0.937 & 5.032 & 0.203 & 0.827 & 0.946 & 0.981 \\
Ours & R & - & - & CD & K & 0.128 & 0.901 & \second{4.898} & 0.198 & 0.834 & 0.948 & 0.983 \\
Ours & R & - & - & CD+K & K & \second{0.125} & \second{0.881} & 4.903 & \second{0.195} & \second{0.840} & \second{0.951} & \second{0.983} \\
\bottomrule
\end{tabular}
\end{table*}

Detailed comparisons can be found in Table~\ref{tab:mono_depth_kitti2015_eigen}. We can observe that our pre-trained models consistently outperforms ImageNet counterparts, as well as randomly initialized models, using either VGG16 or ResNet50 architectures. It is worth noting, however, that converting relative depth to absolute depth is non-trivial. Computing relative depth (\ie, percentile from absolute depth) is a non-linear mapping. The inverse transformation from relative depth to absolute depth is not unique. Following~\cite{zhou17unsupervised}, we multiply our relative depth by a factor as the ratio between relative depth and absolute depth, we get pretty bad results (RMSE of 11.08 vs 4.903), showing our task is non-trivial.

Moreover, pre-training as domain adaptation also improves the performance of our pre-trained model. After fine-tuning our pre-trained model using KITTI's raw videos (Ours CD+K), our ResNet50 model achieves better results than most of the previous methods~\cite{eigen14depth,liu16learning,godard16unsupervised,zhou17unsupervised}. The results are also on par with the state-of-the-art method~\cite{kuznietsov17semi}.


\section{Conclusions and Discussions}
We have proposed a new proxy task for self-supervised learning of visual representations. It requires only access to unlabeled videos taken by a moving camera. Representations are learned by optimizing prediction of relative depth, recovered from estimated motion flow, from individual (single) frames.  Although ostensibly non-semantic, training for this task is likely to encourage the emergence of semantically meaningful representations, since scene understanding can provide significant cues for predicting depth. Indeed, we show this task to be a powerful proxy task, which is competitive with recently proposed alternatives as a means of pre-training representations on unlabeled data. We also demonstrate a novel application of such pre-training, aimed at domain adaptation. When given videos taken by cars driven in cities, self-supervised pre-training primes the downstream urban scene understanding networks, leading to improved accuracy after fine-tuning on a small amount of manually labeled data. 

Our work offers novel insights about one of the most important questions in vision today: how can we leverage unlabeled data, and in particular massive amounts of unlabeled video, to improve recognition systems. While a comprehensive picture of  self-supervision methods and the role they play in this pursuit is yet to emerge, our results suggest that learning to predict relative depth is an important piece of this picture.

While the gap of the performance between self-supervised methods and their ImageNet counterparts is quickly shrinking, none of current self-supervised methods performs better than ImageNet pre-trained models on tasks involving semantics (\eg, semantic segmentation and object detection). This makes pre-training on ImageNet still practically critical for many computer vision tasks. Despite this fact, this does not mean self-supervised methods are unimportant or unnecessary. The value of self-supervised methods lies in the fact that the training data can easily be scaled up without tedious and expensive human effort. 

On other tasks, better performance of self-supervised methods than ImageNet counterparts has been achieved, including our monocular depth estimation and surface normal prediction~\cite{wang17transitive}. Moreover, it has been shown that combining different self-supervised methods can lead to better performance~\cite{doersch17multi,wang17transitive}. All of these make it very promising that representations learned using self-supervised methods may surpass what ImageNet provides us today.

\section*{Acknowledgement}
The experiments were performed using
equipment obtained under a grant from the Collaborative
R\&D Fund managed by the Massachusetts Tech Collaborative.

\clearpage

\bibliographystyle{splncs}
\bibliography{egbib}
\end{document}